\documentclass[lettersize,journal]{IEEEtran}
\usepackage{amsmath,amsfonts}
\usepackage{algorithmic}
\usepackage{algorithm}
\usepackage{array}
\usepackage[caption=false,font=normalsize,labelfont=sf,textfont=sf]{subfig}
\usepackage{textcomp}
\usepackage{stfloats}
\usepackage{url}
\usepackage{verbatim}
\usepackage{graphicx}
\usepackage{cite}
\usepackage{graphicx}
\usepackage{textcomp}
\usepackage{xcolor}
\def\BibTeX{{\rm B\kern-.05em{\sc i\kern-.025em b}\kern-.08em
    T\kern-.1667em\lower.7ex\hbox{E}\kern-.125emX}}
\usepackage{balance} 
\usepackage{bbm}
\usepackage{amsfonts}
\usepackage{epstopdf}
\usepackage{array}
\usepackage{booktabs}
\usepackage{color}
\usepackage{xcolor}
\usepackage{colortbl}
\usepackage{xspace}
\usepackage{multirow}
\usepackage{pifont}
\usepackage{enumitem}
\usepackage{bbding}
\usepackage{fontawesome}
\usepackage{hyperref}
\usepackage{makecell}
\usepackage{microtype}
\usepackage{caption,setspace}
\usepackage{algorithm}
\usepackage{algorithmic}
\usepackage[normalem]{ulem}
\usepackage{hyperref}

\usepackage{amssymb}

\begin{document}

\title{Federated Prototype Graph Learning}

\author{
Zhengyu Wu,
Xunkai Li, 
Yinlin Zhu, 
Rong-Hua Li, 
Guoren Wang, 
Chenghu Zhou

\markboth{Journal of \LaTeX\ Class Files,~Vol.~14, No.~8, August~2021}%
{Shell \MakeLowercase{\textit{et al.}}: FedPG: Prototype-guided Large-scale Federated Graph Learning}}

\maketitle

\begin{abstract}
    In recent years, Federated Graph Learning (FGL) has gained significant attention for its distributed training capabilities in graph-based machine intelligence applications, mitigating data silos while offering a new perspective for privacy-preserve large-scale graph learning.
    However, multi-level FGL heterogeneity presents various client-server collaboration challenges: 
    (1) Model-level: The variation in clients for expected performance and scalability necessitates the deployment of heterogeneous models.
    Unfortunately, most FGL methods rigidly demand identical client models due to the direct model weight aggregation on the server.
    (2) Data-level: The intricate nature of graphs, marked by the entanglement of node profiles and topology, poses an optimization dilemma. 
    This implies that models obtained by federated training struggle to achieve superior performance.
    (3) Communication-level: Some FGL methods attempt to increase message sharing among clients or between clients and the server to improve training, which inevitably leads to high communication costs.
    In this paper, we propose FedPG as a general prototype-guided optimization method for the above multi-level FGL heterogeneity.
    Specifically, on the client side, we integrate multi-level topology-aware prototypes to capture local graph semantics.
    Subsequently, on the server side, leveraging the uploaded prototypes, we employ topology-guided contrastive learning and personalized technology to tailor global prototypes for each client, broadcasting them to improve local training.
    Experiments demonstrate that FedPG outperforms SOTA baselines by an average of 3.57\% in accuracy while reducing communication costs by 168x.

\end{abstract}

\section{Introduction}
\label{sec: introduction}
    \IEEEPARstart{I}{n} recent years, Graph Neural Networks (GNNs) have emerged as a promising approach in the artificial intelligence community, enabling learning and reasoning capabilities over structured data. 
    This growing interest highlights the crucial role of graph learning in advancing machine intelligence, enabling innovation across a wide range of practical applications.
    Specifically, their utility spans a diverse range of domains, including finance financial risk management~\cite{balmaseda2023app_gnn_fina1, hyun2023app_gnn_fina2, qiu2023app_gnn_fina3}, bioinformatics~\cite{bang2023app_gnn_bio1, qu2023app_gnn_bio2, gao2023app_gnn_bio3}, and recommendation systems~\cite{10.1007/s00521-022-07735-y_EHGCN, yang2023app_gnn_rec2, cai2023app_gnn_rec3}.
    Reviewing this promising technology, the core of GNNs is to encode the semantic and structural information of graphs into node embeddings under supervision, thereby integrating insights derived from both node profiles (i.e., node features and labels) and topology. 
    This encoding facilitates the transformation of practical applications into downstream tasks at the node-level~\cite{wu2019sgc, Hu2021ahgae}, edge-level~\cite{cai2021link_prediction2, tan2023link_prediction4}, and graph-level~\cite{zhang2019graph_classification1, yang2022graph_classification3} within the graph machine learning paradigm.

    Despite centralized GNNs have achieved superior efforts in advancing various machine intelligence domains~\cite{arya2024_TPAMI_1, fan2023_TPAMI_2, zheng2023_TPAMI_3, qi2023_TPAMI_4, yin2023_TPAMI_5}, the development of distributed GNN learning methods has become increasingly essential. 
    This shift is driven by the growing demands for privacy, security, and practical applicability in real-world scenarios. 
    For instance, numerous financial institutions and biotechnology companies often restrict data sharing due to privacy concerns but wish to benefit from training with ample data sources~\cite{pan2022fedapp_gnn_fina1, wufederated2023fedapp_gnn_bio2}. 
    Similarly, e-commerce aspires to extract user preferences across multiple platforms for recommendations but prohibits data sharing due to regulatory compliance~\cite{yin2022fedapp_gnn_rec1,mai2023fedapp_gnn_rec3}.
    Meanwhile, as graphs continue to expand, the development of large-scale data intelligence systems capable of billion-scale graph learning is indispensable to effectively analyze these massive datasets, ensuring scalability in real-world applications~\cite{wu2019sgc,frasca2020sign,sun2021sagn,zhang2022pasca,li2024_atp}.
    However, centralized computing and storage settings pose significant limitations in meeting these demands and preserving data privacy~\cite{li2023fedgta,zhang2024feddep,yao2024fedgcn}.
    
\begin{figure*}[t]   
	\centering
    \setlength{\abovecaptionskip}{0.3cm}
    \setlength{\belowcaptionskip}{-0.3cm}
    \includegraphics[width=\linewidth,scale=1.00]{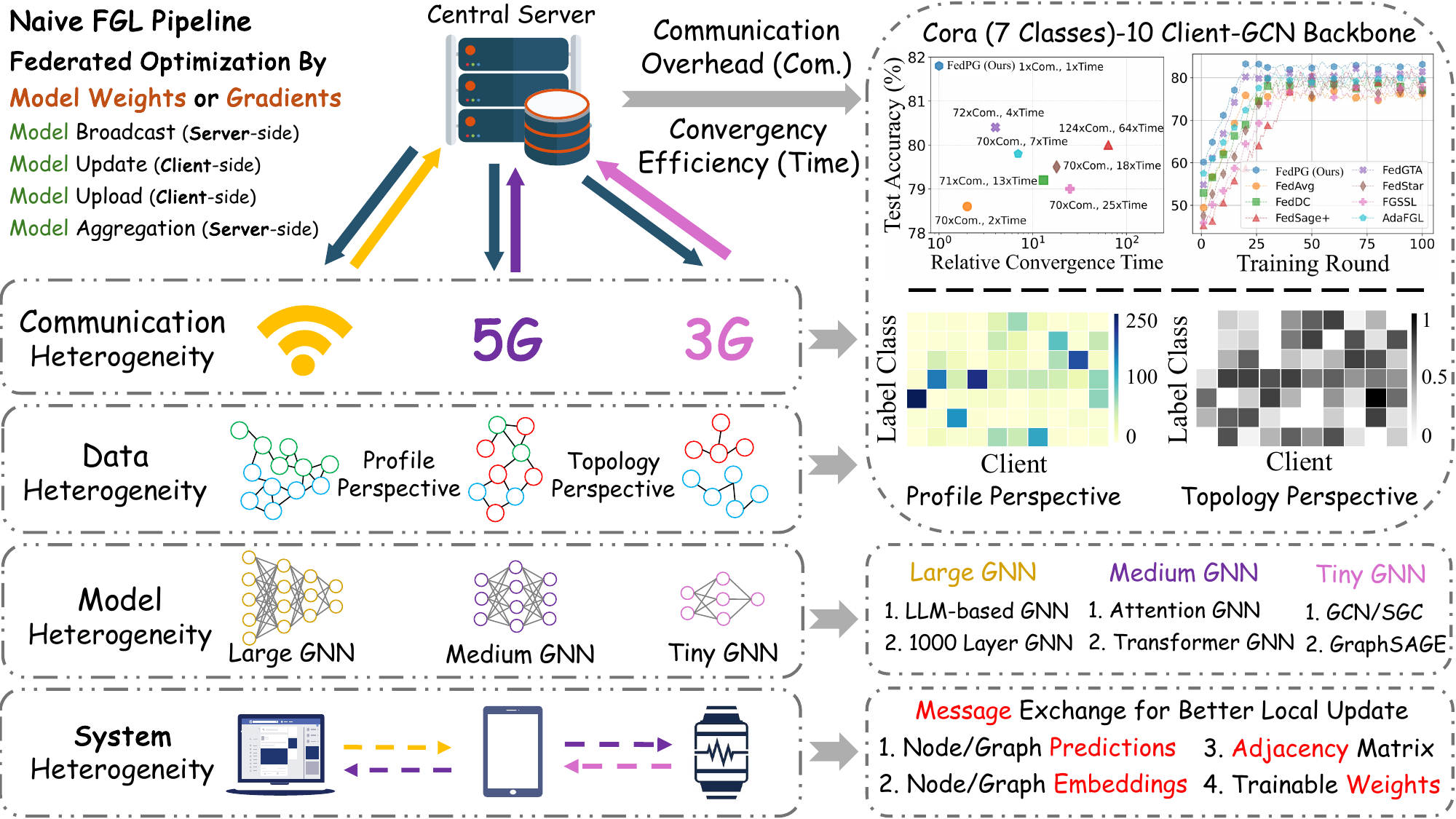}
	\caption{
        A multi-level FGL heterogeneity illustration with the empirical case study.
 }
	\label{fig: heterogeneity_challenges}
\end{figure*}

    Federated Graph Learning (FGL) is a promising technology, which aims to develop collaborative strategies for distributed graph learning across multiple clients and a server without direct data sharing. 
    It is an effective instantiation of distributed machine intelligence, attracting considerable attention in recent years.
    However, considering real-world deployment, there are significant differences among local systems, forming the foundation of the following multi-level FGL heterogeneity that brings challenges for existing methods, as illustrated in Fig.~\ref{fig: heterogeneity_challenges}.

    \textbf{Model Heterogeneity (MH)}: 
    Considering the diverse requirements of local clients in terms of expected performance and varying scale of local data, there exist necessary demands to deploy heterogeneous GNNs (i.e., models with varying amounts of trainable parameters and neural architectures):
    (1) {Large}: LLM-based GNNs~\cite{liu2023ofa} or 1000 layer GNNs~\cite{li2023deepergcn}; 
    (2) {Medium}: attention-based GNNs~\cite{wu2021GraphTrans} or transformer-based GNNs~\cite{chen2022nagphormer}; 
    (3) {Tiny}: GCN~\cite{kipf2016gcn} or GraphSAGE~\cite{hamilton2017graphsage}.
    Therefore, facilitating collaborative training of heterogeneous GNNs across multiple clients is crucial for practical deployments. 
    However, to the best of our knowledge, most FGL methods impose a rigid requirement that demands identical model architectures across clients, as they directly aggregate model weights on the server.

    \textbf{Data Heterogeneity (DH)}: 
    Compared to the text or images in Euclidean space, graphs inherently exhibit complexity due to the entanglement of node profiles (i.e., node features and labels) and topology.
    To illustrate this issue thoroughly, we examine Subgraph-FL as an instance of FGL with the local semi-supervised node classification, exploring the unique graph-based data Non-iid challenges in Fig.~\ref{fig: heterogeneity_challenges}. 
    (1) {Profile Perspective}: 
    Due to the well-known graph homophily assumption~\cite{mcpherson200homophily_theory1, M2003Mixing_homophily_theory2, 0Networks_homophily_theory3}, nodes with similar feature distributions or same labels tend to form communities collected by the different clients, causing significant disparities in the label distribution heat map, where the value denotes the number of nodes belongs to each client. 
    (2) {Topology Perspective}: 
    We visualize the connection patterns of each client to illustrate the diverse topology. 
    Specifically, we calculate node homophily~\cite{pei2020geomgcn} for each label class to generate this heat map, where darker colors indicate that the neighbors of nodes in a given class predominantly belong to the same class.
    Notably, in Graph-FL represented by local graph classification, topology heterogeneity can be examined through topology statistics (e.g., degree distribution and largest component size), as highlighted in recent studies~\cite{xie2021gcfl, tan2023fedstar}. 
    
    \textbf{Communication Heterogeneity (CH)}: 
    To achieve efficient federated training, minimizing communication costs is crucial for avoiding deployment impediments caused by communication overhead.
    However, most FGL studies~\cite{zhang2021fedsage, wu2021fedgnn, zhao2022fedgsl, baek2022fedpub, chen2024fedgl} overly rely on message sharing to address {DH}. 
    This implies that existing approaches tend to sacrifice running efficiency (Communication Cost and Time Cost) to enhance predictive performance (Highest Accuracy in Convergence Curve).
    To further illustrate, we provide empirical analysis to reveal the limitations of existing studies by visualizing their convergence curve and running efficiency in Fig.~\ref{fig: heterogeneity_challenges}.
    Notably, existing FGL methods often incorporate message exchange mechanisms between distinct local devices to capture potential dependencies within multi-client local graphs in addition to client-server communication~\cite{zhang2021fedsage, zhao2022fedgsl, yao2024fedgcn, zhang2024feddep}. 
    Based on this, compared to the most competitive baselines, our proposed FedPG achieves SOTA performance and exhibits the fastest convergence while reducing communication overhead by two orders of magnitude.

    To address the above multi-level FGL heterogeneity, Federated Prototype Learning (FPL) is a promising technology, which has been widely applied in the Computer Vision (CV) domain in recent years~\cite{tan2022fedproto, dai2023fednh, huang2023fpl, liao2023hyperfed, zhang2024fedtgp}.
    Unlike directly uploading models or gradients, the multi-client collaboration pipeline of FPL is as follows:
    {Step 1}: Each client generates label class-wise prototypes based on the class-specific data sample embeddings; 
    {Step 2}: These local prototypes are uploaded to the server to generate global prototypes; 
    {Step 3}: These global prototypes are broadcast to clients to guide local training. 
    The key innovation lies in replacing model weights or gradients with prototypes as the shared information carrier during collaborative training.

    Despite the effectiveness of FPL in CV, the unique {DH} in FGL hinders its direct application, resulting in inferior predictions shown in Sec.~\ref{sec: Performance Comparison}.
    Therefore, we introduce FedPG as the first graph-based general FPL, enabling model-agnostic federated optimization (handle {MH}). 
    Specifically, on the {client}: FedPG introduces a topology-aware prototype encoding strategy that integrates the elaborated graph context, which refers to a general characterization of semantic and structural insights.
    On the {server}: FedPG utilizes the uploaded prototypes to achieve topology-guided contrastive learning (CL).
    This approach facilitates the trainable generation of global prototypes shared among participating clients, adaptively extracting global consensus. 
    After that, personalized techniques are employed to customize global prototypes for each client, empowering local training (handle {DH}). 
    Notably, in terms of algorithm complexity, the embeddings used to generate prototypes are much smaller than the multiple layers of trainable neurons, resulting in a significant reduction of communication costs (handle {CH}).
    More details are shown in Sec.~\ref{appendix: FedPG Algorithm and Complexity Analysis} and Sec.~\ref{sec: Performance Comparison}.
    
    We note that the recently proposed FGGP~\cite{wan2024fggp} also introduces prototype learning in FGL. 
    However, its primary aim is to address the domain shift challenge through well-designed local training by prototypes, setting it apart from our proposed FedPG. 
    A thorough examination of this disparity will be provided in Sec.~\ref{sec: Performance Comparison}.
    As for privacy concerns about prototypes, we believe that prototypes, serving as high-level statistic information about node embeddings, are distinguished from original raw data profiles.
    This distinction obviously makes the attack harder~\cite{wang2024prototype_privacy1,wang2023prototype_privacy2,zuo2024prototype_privacy2} and increases deployment robustness. 
    In addition, we propose noise-based protection mechanisms to further mitigate privacy concerns.
    More implementation details and experimental results can be found in Sec.~\ref{sec: Robustness in Noise-based Privacy Preservation}.

    \textbf{Our contributions.}
     (1) \underline{\textit{New Perspective.}}
     Motivated by the challenges of FGL deployment and practical distributed machine intelligence, we present a brief review and valuable case study of multi-level FGL heterogeneity in Fig.~\ref{fig: heterogeneity_challenges}. 
     This investigation delineates three distinct challenges that current methods struggle to address.
     (2) \underline{\textit{New Method.}}
     Drawing inspiration from FPL, we propose FedPG as the first model-agnostic general optimization framework for multi-level FGL heterogeneity. 
     On the client, the core of FedPG is to generate prototypes that thoroughly extract the local graph context.
     On the server, FedPG aims to adaptively customize the optimal global prototypes for each client to provide better local training guidance.
     (3) \underline{\textit{SOTA Performance.}}
     Extensive experiments conducted on 14 datasets with different scales (up to billion-level) demonstrate that FedPG not only achieves SOTA performance but also offers high training efficiency and low communication costs.
     Specifically, FedPG outperforms the most representative FedSage+~\cite{zhang2021fedsage} by 3.2\%-5.7\% in accuracy, running 25x-148x faster, and reducing 39x-1564x communication overhead.

\section{Preliminaries}
\label{sec: preliminaries}
    Considering the various graph-based downstream tasks in FGL (e.g., node classification, link prediction, and graph classification), we introduce FedPG using the node classification within Subgraph-FL paradigm depicted in Fig.~\ref{fig: heterogeneity_challenges} for reader-friendly and clarity.
    Further details about the generalization of FedPG in Graph-FL can be found in Sec.~\ref{appendix: Experiments for Graph-FL}.
    Now, we present the formalization of the problem and review recent related work to establish a clear research context for our approach.
    
\textbf{Problem Formalization.}
    In Subgraph-FL, each client manages a local graph learning model (i.e., GNNs with varying architectures) and a private subgraph from a global graph perspective. 
    They aim to collaboratively improve the performance of the local semi-supervised node classification.
    Formally, we consider $N$ clients possessing locally collected data, denoted as $\left\{\mathcal{G}_1,\dots,\mathcal{G}_N\right\}$. 
    Here, $\mathcal{G} = (\mathcal{V}, \mathcal{E})$ comprises $|\mathcal{V}|=n$ nodes and $|\mathcal{E}|=m$ edges, with an adjacency matrix (including self-loops) $\hat{\mathbf{A}}\in\mathbb{R}^{n\times n}$. 
    The feature matrix is $\mathbf{X} = \{x_1,\dots,x_n\}$, where $x_v$ is the feature vector of node $v$.
    Besides, $\mathbf{Y} = \{y_1,\dots,y_n\}$ is the label matrix, where $y_v$ is a one-hot vector and $|\mathcal{C}|$ is the class number.
    Notably, clients may have varying $n$ and $m$, but share the same dimension of features and labels.
    The semi-supervised node classification task is based on the topology of labeled set $\mathcal{V}_L$ and unlabeled set $\mathcal{V}_U$, and the nodes in $\mathcal{V}_U$ are predicted based on the model supervised by $\mathcal{V}_L$.
    
    To achieve multi-client collaboration, FGL typically follows the standard federated training pipeline shown in Fig.~\ref{fig: heterogeneity_challenges}.
    For instance, FedAvg~\cite{mcmahan2017fedavg} is the most widely used optimization strategy for multi-client collaboration. 
    The key idea of this method is to implement model aggregation by using a simple weighted average of the model parameters received from each participating client. 
    The weights used in the aggregation for each client are proportional to its local data size. 

\textbf{Federated Learning.}
    Despite FedAvg offering a simple yet effective strategy for distributed training, the inherent system heterogeneity poses significant challenges.
    Consequently, FedProx~\cite{li2020fedprox} improves local updates based on the global model to address DH and dynamically adjusts iteration rounds on local devices for CH. 
    Scaffold~\cite{karimireddy2020scaffold}, MOON~\cite{li2021moon}, and FedDC~\cite{gao2022feddc} further address DH through controlled variables, local model-wise CL, and drift variables, respectively.
    Despite the effectiveness of these optimization methods, they lack a general solution for all above heterogeneity issues. 
    Fortunately, FedProto~\cite{tan2022fedproto} introduces FPL to address these challenges simultaneously, establishing a standard prototype collaboration pipeline. 
    Based on this, FedNH~\cite{dai2023fednh}, FPL~\cite{huang2023fpl}, and FedTGP~\cite{zhang2024fedtgp} further optimize FedProto through separable prototype initialization, local CL, and trainable prototype correction, respectively.

\textbf{Federated Graph Learning.}
    Considering the diverse nature of graph-based downstream tasks, existing FGL studies can be broadly categorized into:
    (1) {Graph-FL}: 
    In this setting, each client collects locally private graphs, aiming to collaboratively address local graph-level downstream tasks (e.g., graph classification and regression). 
    Current researches underscore the structure Non-iid problem, which is marked by notable variations in graph structural statistics, including degree distribution or average shortest path length.
    Building upon this, GCFL+~\cite{xie2021gcfl} and FedStar~\cite{tan2023fedstar} apply personalized techniques and graph structure encoder knowledge sharing to improve performance.
    (2) {Subgraph-FL}: As emphasized in Fig.~\ref{fig: heterogeneity_challenges} and Sec.~\ref{sec: introduction}, significant optimization challenges arise due to the profile and topology heterogeneity. 
    To improve node classification performance, existing studies can be categorized into two types: 
    (i) sharing additional messages among clients or between clients and the server to promote federated collaborative optimization, including FedGNN~\cite{wu2021fedgnn}, FedSage+~\cite{zhang2021fedsage}, FedGSL~\cite{zhao2022fedgsl}, Fed-PUB~\cite{baek2022fedpub},
    FedGTA~\cite{li2023fedgta},
    FedGT~\cite{zhang2024fedgt}, FedGL~\cite{chen2024fedgl}, and FedDEP~\cite{zhang2024feddep}; 
    (ii) applying personalized techniques on the server or client to fit local subgraph, including FedLit~\cite{xie2023fedlit}, FGSSL~\cite{huang2023fgssl}, GCFGAE~\cite{guo2023GCFGAE}, AdaFGL~\cite{li2024adafgl}, and FGGP~\cite{wan2024fggp}.

\section{FedPG Framework}
\label{sec: federated prototype graph learning}

\begin{figure*}[t]   
    \setlength{\abovecaptionskip}{0.25cm}
    \setlength{\belowcaptionskip}{-0.25cm}
	\centering
 \includegraphics[width=\linewidth,scale=1.00]{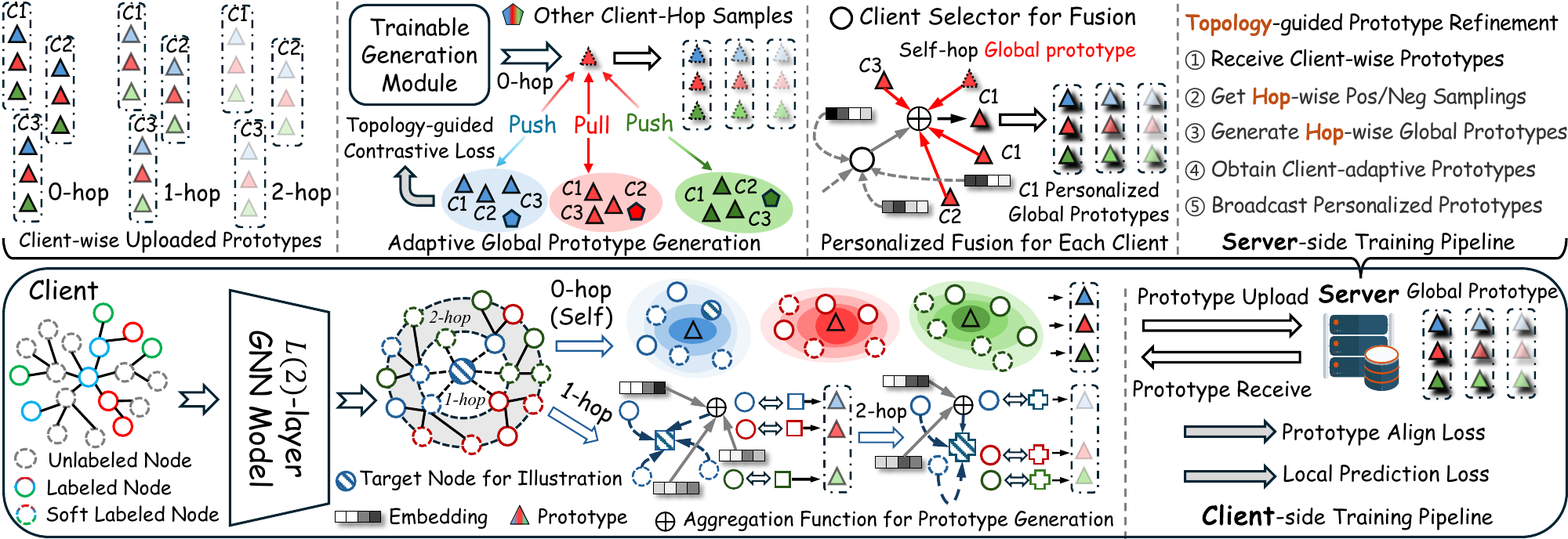}
	\caption{
    Overview of our proposed FedPG framework. 
 }
	\label{fig: framework}
\end{figure*}

    In this section, we introduce FedPG as a general optimization framework for multi-level FGL heterogeneity.
    In contrast to the CV-based FPL, the key challenges of FedPG lie in effectively managing node correlations induced by topology, while converting these correlations for optimal prototypes to drive FGL.
    Specifically, on the \textbf{Client-side}, FedPG first harnesses node embeddings produced by local GNN to capture the interplay between profiles and topology. 
    Subsequently, by leveraging explicit topology, FedPG adaptively aggregates class-annotated node embeddings to derive topology-aware multi-level prototypes.
    \textit{Motivation: this strategy guarantees a holistic extraction of local private graph semantic context.}
    On the \textbf{Server-side}, FedPG first utilizes uploaded prototypes to derive positive and negative samples by label classes and topological guidance. 
    Then, FedPG formulates a well-designed CL loss using these data samples, supervising the trainable module to generate universal global prototypes shared among participating clients in each round.
    \textit{Motivation: this strategy extracts global consensus for each label class and further refines the semantic generalization of the generated prototypes within topological guidance.}
    After that, FedPG employs a personalized strategy to tailor the local prototypes broadcast to each client.
    \textit{Motivation: this strategy aims to ensure fast and stable local convergence for each client under personalized prototype supervision, thus maximizing training efficiency and accommodating granularity variations among clients}.
    The detailed algorithm overview of FedPG is shown in Fig.~\ref{fig: framework} and Sec.~\ref{appendix: FedPG Algorithm and Complexity Analysis}.
    
    In a nutshell, FedPG thoughtfully leverages the unique characteristics of graph data (i.e., client-side topology-aware prototype representation and server-side topology-guided prototype refinement) to achieve graph-based FPL, focusing on practicality rather than complex theories and technologies.
    This practicality is demonstrated by its consistent SOTA performance across diverse tasks and its ability to address FGL heterogeneity.
    Given its strong empirical performance, efficiency, and flexibility, we emphasize that FedPG serves as a simple baseline for the FGL community, providing a valuable inspiration for future prototype-based FGL studies.

\subsection{Client-side Local Prototype Representation}
\label{sec: Topology-aware Local Prototype Representation}
    During our investigation, most FPL methods~\cite{tan2022fedproto,dai2023fednh,zhang2024fedtgp,huang2023fpl,wan2024fggp} adhere to a common approach to generating local prototype $\mathcal{P}$. 
    Specifically, they directly aggregate the embeddings of sample $i$ corresponding to each class $c$ as follows:
\begin{equation}
\label{eq: naive local prototype generation}
    \begin{aligned}
    \mathcal{P}_c^k=\frac{1}{\left|\mathcal{S}_c^k\right|} \sum_{\left(x_i, y_i\right) \in \mathcal{S}_c^k} f\left(x_i\right),
    \end{aligned}
\end{equation}
    where $\mathcal{S}_c^k$ is the set of samples annotated with class $c$ in client $k$. 
    Notably, model $\mathcal{M}$ typically comprises embedding and prediction components $f$ and $g$. 
    The former maps initial features to an embedding space for semantic extraction, while the latter produces task-specific outputs (e.g., soft labels).
    As shown in Eq.~(\ref{eq: naive local prototype generation}), local prototypes are derived from class-specific embedding averaging, which encapsulates high-level statistic-conveyed semantic information.

    Despite its effectiveness, directly applying Eq.~(\ref{eq: naive local prototype generation}) in FGL has limitations. 
    Unlike the relatively independent images, graph nodes exhibit intricate dependencies. 
    Consequently, naive class-specific embedding averaging fails to capture the nuanced attributes of local graphs, disregarding implicit connections among label classes.
    Therefore, we are motivated to define a topology-aware prototype representation as follows:

\begin{equation}
\label{eq: fedpg local prototype generation}
    \begin{aligned}
    &\left|\mathcal{N}_c^h(v_i)\right|\!=\!L_{c}^h,\;\mathcal{P}_{c,h}^k\!=\!\frac{1}{\left|\mathcal{S}_{c}^k\right|}\!\!\sum_{\left(x_i, y_i\right) \in \mathcal{S}_c^k}\sum_{v_j\in\mathcal{N}_c^h(v_i)}\!\!\!\!\frac{w_{j}f\left(x_j\right)}{{L_{c}^h}},\\
    &w_{j} \!=\! e^{\delta\left(\mathbf{E}_j^{\left(l\right)}\right)} \!/\! \sum_{k=1}^{K}e^{\delta\left(\mathbf{E}_j^{\left(k\right)}\right)},\mathbf{E}_j^{(l)} \!\!=\! \operatorname{MLP}\!\left(x_1||\cdots||x_{L_{c}^h}\right),
    \end{aligned}
\end{equation}
    where $\mathcal{N}_c^h(v_i)$ is the $h$-hop neighbors of current node $i$ annotated with label class $c$ and $w_{j}$ denotes trainable parameters, used to adaptively aggregate sample embeddings from $\mathcal{N}_c^h(v_i)$ when generating prototypes.
    The key insight stems from the prevalent message-passing mechanisms~\cite{xu2018jknet}. 
    This means that $L$-layer GNNs learn node representations by integrating information from $L$-hop neighborhoods.
    Expanding on this notion, the collection of nodes within the $L$-hop neighborhood of a specific node is defined as its Receptive Field (RF).
    Hence, to capture the meaningful interaction between profiles and topology, we adaptively aggregate the embeddings of nodes by their respective RF and attention $w$. 
    This aggregation provides a multi-granularity representation for topology-aware prototypes.
    Meanwhile, we set $h=\{0,\dots, \min(L)\}$ to reduce manual overheads, where $\min(L)$ is the minimum local model layers across all clients and $h=0$ signifies that only consider node itself (equivalent to Eq.~(\ref{eq: naive local prototype generation})).
    To this end, we have obtained general prototypes for each label class $[\mathcal{P}_{c,0},\dots,\mathcal{P}_{c,L}]$ within RF prompted.
    These prototypes provide a comprehensive and high-quality extraction of local graph context.
    Notably, this accomplishment is attributed to the attention mechanisms and thorough utilization of node semantic and structural insights.

\subsection{Server-side Global Prototype Refinement}
\label{sec: Adaptive Global Prototype Generation}
    Upon receipt of prototypes uploaded by participating clients in each round, the server strives to process and broadcast the most suitable global prototypes to each client for supervision.
    This is to maximize local training efficiency to achieve more stable and faster convergence.
    Reviewing existing FPL methods, the most widely used approach for global prototype generation is the naive weight-average aggregation proposed by~\cite{tan2022fedproto}: 
    \begin{equation}
\label{eq: naive global prototype generation}
    \overline{\mathcal{P}}_{c}^t=\frac{1}{\left|{N}^t_c\right|}\sum_{i\in{N}^t_c} \frac{n_c^i}{n_c^1+\dots+n_c^{\left|{N}^t_c\right|}}\mathcal{P}_{c}^i,
\end{equation}
    where $\overline{\mathcal{P}}_{c}^t$ represents the server-side $c$-class prototype obtained at round $t$, $N_c^t$ denotes the set of clients participating uploading $c$-class prototypes, and $n_c^i$ denotes $c$-class data size on the $i$ client.
    Although this strategy appears simple and intuitive, the unique data heterogeneity in the entanglement of profiles and topology often leads to substantial variations in local graph prototypes. 
    Naively aggregating them results in semantic confusion and neglects the RF prompts derived from $\mathcal{P}_{c,h}$, thus adversely affecting multi-client collaboration. 
    
    Despite recent progress in global prototype generation~\cite{dai2023fednh, huang2023fpl, zhang2024fedtgp}, most of these methods are primarily tailored for CV.
    Consequently, they overlook the crucial graph context driven by topology, resulting in sub-optimal performance, as shown in Sec.~\ref{sec: Performance Comparison}. 
    Besides, the recently proposed FPPG focuses solely on the client-side domain shift correction and only achieves client-side graph-based prototype refinement, neglecting global prototypes. 
    This motivates us to propose a comprehensive and adaptive prototype generation strategy for FGL (i.e., universal global prototype and then personalized global prototype) while aligning with the topology-aware prototypes introduced in Sec.~\ref{sec: Topology-aware Local Prototype Representation}, through the following two steps.

\textbf{Universal Global Prototype Generation.}
    Essentially, server-side prototype collaboration resembles the model aggregation in FedAvg. 
    The primary target is to integrate local knowledge on the server to attain global consensus. 
    Then, the server broadcasts this consensus to supervise local updates.
    To break the limitations of existing FPL methods in perceiving topology-driven graph context, we introduce a trainable Global Prototype Generator (GPG) supervised by a topology-guided CL loss under RF prompts.
    The aim is to adaptively extract the optimal global consensus from all participating clients to generate universal prototypes. 
    We expect them to exhibit universal generalization to topology-driven $c$-class global semantics and maintain semantic separability among different label classes. 
    The above process can be formally defined as:
\begin{equation}
\small
\label{eq: fedpg server loss}
    \begin{aligned}
&\;\;\;\;\;\;\;\;\;\;\;\;\min_{\tilde{\mathcal{P}}_{c,h}}\sum_{c=1}^{|\mathcal{C}|}\sum_{h=0}^{L}\mathcal{L}_{c,h},\;\tilde{\mathcal{P}}_{c,h} = {\operatorname{GPG}}\left(\ddot{\mathcal{P}}_{c,h}, \theta\right),\\
    &\tilde{D}=D\left(\tilde{\mathcal{P}}_{c,h},\mathcal{P}_i\right)=\frac{\tilde{\mathcal{P}}_{c,h}\cdot\mathcal{P}_i}{\Vert\tilde{\mathcal{P}}_{c,h}\Vert\times\Vert\mathcal{P}_i\Vert},\;\mathcal{Q}_i(c)=\sum_{i\in\mathcal{S}}\mathcal{P}_i/|\mathcal{S}|,\\
    &\tilde{M}\!=\!M\left(\mathcal{Q}_i(c),\mathcal{Q}_i(c\prime)\right)\!\!=\!\min\!\left(\!\max_{c, c\prime\in\mathcal{C},c\neq c\prime}\!\!\!D\!\left(\mathcal{Q}_i(c),\mathcal{Q}_i(c\prime)\right)\!,\!\epsilon\!\right),\\
    &\mathcal{L}_{c,h}\!=\!-\!\log\frac{\sum_{i\in\mathcal{S}_{c,\vartriangle \!h}}\!\!\exp{\left(\tilde{D}+\tilde{M}\right)}}{\sum_{i\in\mathcal{S}_{c,\vartriangle\!h}}\!\!\exp\!{\left(\tilde{D}+\tilde{M}\right)} + \sum_{i\in\mathcal{S}_{c\prime,\vartriangle \!h}}\!\!\exp\!{\left(\tilde{M}\right)}},
    \end{aligned}
\end{equation}
    where $\tilde{\mathcal{P}}_{c,h}$ and $\ddot{\mathcal{P}}_{c,h}$ are the generated universal global prototypes and trainable prototypes initialized randomly within $h$-hop. 
    In our implementation, the trainable module $\operatorname{GPG}(\cdot)$ consists of two fully connected layers with a ReLU activation function. 
    The generated class-specific and hop-specific universal global prototypes share the same $\operatorname{GPG}$ weights $\theta$. 
    
    In our proposed topology-guided CL loss $\mathcal{L}_{c,h}$ for universal global prototype generation, $D(\cdot)$ is the distance metric between the query prototype $\mathcal{P}_i$ and trainable anchor prototype $\tilde{\mathcal{P}}_{c,h}$, $\mathcal{Q}_i(\cdot)$ is the embedding center of the $i$-th query prototypes, $M(\cdot)$ is the margin function based on the adaptive maximum discrepancy between prototypes of different classes and hops, and $\mathcal{S}_{c,h}$ is the set of samples annotated with class c within $h$-hop and there exists $\vartriangle\!\!h\in\{0,\dots,\min(L)\},\vartriangle\!h\neq h$.
    Notably, $\epsilon$ is a hyperparameter employed to prevent the margin from growing infinitely.
    The key intuitions in Eq.~(\ref{eq: fedpg server loss}) are as follows.
    
    \underline{RF-prompted Context}.
    The client-wise trainable multi-granularity prototypes $\{\tilde{\mathcal{P}}_{c,0},\dots,\tilde{\mathcal{P}}_{c,L}\}$ for each label class under different hops enhance local graph context extraction, aiding in the discovery of more effective global consensus from a topological perspective. 
    Meanwhile, $\vartriangle\!h$ achieves a topology-driven sampling to obtain abundant query sets, providing more informative objectives and maximizing the benefits of subsequent CL.
    In a nutshell, this approach enables topology-driven, fine-grained graph prototype pattern discovery, laying a solid foundation for subsequent intelligent analysis.
    
    \underline{Class-oriented CL}.
    Based on the query samples, for each class, we pull closely the anchor prototype (i.e., generated universal global prototype) and same-class query prototypes while distancing the anchor prototype from query prototypes of different classes. 
    Through CL supervision formulated by topology-driven RF at different levels, the trainable module generates universal global prototypes containing real graph semantic knowledge for each label class.

    \underline{Margin-based Enhancement}.
    Although RF-prompted Context and Class-oriented CL aid in generating universal global prototypes with topological consensus and intra-class semantic consistency, the learned inter-class separation boundaries may still lack clarity. 
    Hence, we integrate quantified adaptive maximum discrepancy into the CL loss to enforce explicit inter-class separation during learning.
    Building upon Class-oriented CL, this strategy further effectively increases distances among anchor prototypes with different classes.

\textbf{Personalized Global Prototype Fusion.}
    Despite the robust generalization capabilities of the generated universal prototypes, broadcasting the same prototype to all participating clients inevitably compromises the ability to fit the local data.
    Especially in the DH, fine-grained graph context differences among multiple clients cannot be ignored. 
    Therefore, it motivates us to propose personalized techniques, tailoring the most suitable global prototypes for each participating client based on the generated universal prototypes.
    This strategy aims to achieve faster and more stable local convergence, along with higher accuracy.
    Formally, this personalized process is defined as:
\begin{equation}
\label{eq: fedpg personalized prototype fusion}
    \begin{aligned}
    &\;\;\;\;\;\;\;\;\dot{\mathcal{S}}_i =\{k|\mathrm{sim}(i,k)\ge \lambda\},\forall i,k\in N_c^t,\\
    &\dot{\mathcal{P}}_{c,h}^i = \alpha\tilde{\mathcal{P}}_{c,h}+(1-\alpha)\sum_{j\in\dot{\mathcal{S}}_i}\frac{n_j}{n_j+\dots+n_{|\dot{\mathcal{S}}_i|}}\mathcal{P}_j,
    \end{aligned}
\end{equation}
    where $\dot{\mathcal{P}}_{c,h}^i$ is the personalized prototype sent to client $i$, $\dot{\mathcal{S}}_i$ is the prototype fusion set for client $i$. 
    $\mathrm{sim}(\cdot)$ is the similarity function employed to quantify differences in local graph context and $\lambda$ is the threshold. 
    In our implementation, we default to using cosine similarity.
    This strategy aims to quantify fine-grained semantic differences among clients, facilitating personalized fusion among similar clients. 
    Intuitively, personalized global prototypes for each client integrate the most suitable global semantics from their local private graphs, thereby maximizing local training efficiency.
    Building upon this, the $\mu$-based flexible prototype-assisted local training loss is defined as $\mathcal{L} = \mathcal{L}_{ce} + \mu\mathcal{L}_{proto}$.
    Specifically, given $k$-client prediction $\hat{\mathbf{Y}}$, the cross entropy and prototype loss are as follows:
\begin{equation}
\label{eq: fedpg local loss}
    \begin{aligned}
    &\mathcal{L}_{ce}^k = \sum_{i \in \mathcal{V}_L^k} \sum_j \mathbf{Y}_{i j} \log \left(\operatorname{softmax}(\hat{\mathbf{Y}})_{i j}\right),\\
    &\;\;\;\;\;\mathcal{L}_{proto}^k= \sum_{c=1}^{|\mathcal{C}|}\sum_{h=0}^L\Vert \mathcal{P}_{c,h}^k-\dot{\mathcal{P}}_{c,h}^k \Vert_F.\\
    \end{aligned}
\end{equation}

    To this end, universal global prototypes and personalized global prototypes not only inherit specific label class knowledge from each client but also further extract real semantics for each label class from a topological perspective. 
    They demonstrate compact semantic consistency within the same classes and separable semantic boundaries within different classes, thereby exhibiting robustness in multi-level FGL heterogeneity.

\subsection{Additional Optimization Details in $\operatorname{GPG}$}
\label{appendix: FedPG Server-side Optimization Direction}
    In FedPG, we introduce topology-guided prototype CL on the server, enabling the adaptive generation of universal global prototypes. 
    Intuitively, we aim for these prototypes to inherit private graph semantic context from each local prototype within RF and extract the real semantics of each label class, exhibiting consistent intra-class semantics and clear inter-class boundaries. 
    
    In our implementation, we contrast the anchor prototypes generated by the trainable module with prototypes from each query sample sharing the same class and prototypes from other classes. 
    Specifically, we aim to minimize the distance between anchor prototypes and class semantics-aligned prototypes. 
    Simultaneously, we maximize the distance from anchor prototypes to others.
    This CL strategy ensures compact intra-class semantic consistency and clear inter-class decision boundaries in FGL from a topology perspective.
    The following optimization objectives can be derived:
\begin{equation}\small
\label{eq: derived fedpg server loss}
    \begin{aligned}
    \mathcal{L}_{c,h}&=\!-\log\frac{\sum_{i\in\mathcal{S}_{c,\vartriangle h}}\!\exp{\left(\tilde{D}+\tilde{M}\right)}}{\sum_{i\in\mathcal{S}_{c,\vartriangle\!h}}\!\exp{\left(\tilde{D}+\tilde{M}\right)} \!+\! \sum_{i\in\mathcal{S}_{c\prime,\vartriangle h}}\!\exp{\left(\tilde{M}\right)}}\\
    &=\!\log\frac{\sum_{i\in\mathcal{S}_{c,\vartriangle\!h}}\!\exp{\left(\tilde{D}\!+\!\tilde{M}\right)} + \sum_{i\in\mathcal{S}_{c\prime,\vartriangle\!h}}\!\exp{\left(\tilde{D}\right)}}{\sum_{i\in\mathcal{S}_{c,\vartriangle\!h}}\!\exp{\left(\tilde{D}+\tilde{M}\right)}}\\
    &=\!\log\left(1\!+\!\frac{\sum_{i\in\mathcal{S}_{c\prime,\vartriangle\!h}}\exp{\left(\tilde{D}\right)}}{\sum_{i\in\mathcal{S}_{c,\vartriangle\!h}}\exp{\left(\tilde{D}+\tilde{M}\right)}}\right)\\
    &=\!\underbrace{\log\!\!\left(\sum_{i\in\mathcal{S}_{c\prime,\vartriangle\!h}}\!\exp{\left(\tilde{D}\right)}\!\!\right)}_{\textbf{Semantic}\; \textbf{Consistency}} \!-\! \underbrace{\log\!\!\left(\sum_{i\in\mathcal{S}_{c,\vartriangle\!h}}\!\exp{\left(\tilde{D}+\tilde{M}\right)}\!\!\right)}_{\textbf{Consistency}\; + \;\textbf{Separability}}.
    \end{aligned}
\end{equation}
    In Eq.~(\ref{eq: derived fedpg server loss}), we reformulate the topology-guided CL loss for a detailed analysis of its optimization directions. 
    Based on our analysis, we derive the following key insights of generated universal global prototypes regarding its underlying principles:
    
    (1) Intra-class semantic consistency:
    For CL, minimizing Eq.~(\ref{eq: derived fedpg server loss}) implies minimizing $\log\!\left(\!\sum_{i\in\mathcal{S}_{c\prime,\vartriangle\!h}}\!\exp\!{\left(\!\tilde{D}\right)}\!\right)$. 
    Since the logarithmic function with a base greater than 1 and $\tilde{D}(\cdot)\!\in\![0,1]$, this encourages $\tilde{D}$ to approach 1, effectively maximizing the distance between anchor prototypes and negative samples. 
    Similarly, maximizing the objective drives $\log\!\left(\!\sum_{i\in\mathcal{S}_{c,\vartriangle\!h}}\!\exp\!{\left(\!\tilde{D}\right)\!}\right)$ to approach 0, effectively minimizing the distance between anchor prototypes and positive samples. 
    These objectives ensure the semantic accuracy of the generated anchor prototypes, manifested in intra-class semantic consistency.
    
    (2) Inter-class semantic boundaries separability: 
    While (1) guarantees accurate extraction of semantic information for each label class, the separation boundaries between anchor prototypes representing different classes may not be sufficiently clear, especially when generating prototypes for each hop.
    This is due to without distinguishing the semantic boundaries, graph homophily brings semantic space distortions, leading to the confusion of label class semantics. 
    Hence, to ensure discriminative semantic spaces for prototypes, we introduce adaptive maximum discrepancy to increase the optimization difficulty of $\log\!\left(\!\sum_{i\in\mathcal{S}_{c,\vartriangle h}}\!\exp{\!\left(\!\tilde{D}\right)\!}\right)$. 
    This allows the trainable module to focus more on aligning the semantic representations between anchor prototypes and positive samples, thereby increasing the representation distance between anchor prototypes of different classes and achieving separable inter-class semantic boundaries.

\subsection{FedPG Algorithm and Complexity Analysis}
\label{appendix: FedPG Algorithm and Complexity Analysis}
    To this end, we have provided the motivations and computation details of our approach.
    In this section, for a more comprehensive presentation, we provide the complete algorithm of FedPG in Algorithm~\ref{alg: fedpg_client}-\ref{alg: fedpg_server}.
    Based on this, we provide a clear motivation behind our proposed FedPG and reveal its algorithm complexity advantages compared to prevalent FL and FGL studies.
    Specifically, we first conduct a comprehensive review in Table~\ref{tab: review fgl studies}, where Subgraph/Graph-FL denotes the evaluation scenario of these methods within their respective original paper. 
    Notably, most of them are confined to a single FGL scenario, lacking descriptions of their adaptability to other scenarios. 
    In our investigation, we reasonably extend these methods to encompass broader FGL scenarios. 
    Regrettably, they all fail to demonstrate competitive performance in the new FGL scenario. 
    In contrast, we present a detailed description in Sec.~\ref{appendix: Experiments for Graph-FL} that naturally extends FedPG to Graph-FL. 
    Experimental results presented in Sec.~\ref{sec: Performance Comparison} and Sec.~\ref{appendix: Experiments for Graph-FL} validate the satisfied generalization of FedPG.
    Moreover, Table~\ref{tab: review fgl studies} underscores that existing methods often present significant communication overhead. 
    For instance, while FedSage+ and FedGTA achieve competitive performance, they necessitate the exchange of other privacy-sensitive information. 
    Although other methods exhibit relatively lower communication overhead, they still lag behind FedPG in terms of accuracy and effectiveness.
    This review underscores the limitations of existing approaches in addressing the multi-level heterogeneity inherent in the FGL.

\begin{algorithm}[h]
\caption{Prototype-guided FGL (FedPG-Client)} 
\label{alg: fedpg_client}
\begin{algorithmic}[1] 
    \FOR { each participating client $i = 1, ..., N_{t}$ in round $t$} 
       \FOR {each local model training epoch $e = 1, ..., E$}
        \STATE Obtain the local prediction for nodes in the training sets;\\
        \IF {$t=1$}     
            \STATE Update client-independent GNN model weights $f$ and $g$ only by cross-entropy loss;
        \ELSE
            \STATE \uwave{\textit{Receive the personalized global prototype}};
            \STATE  Update client-independent GNN model weights $f$ and $g$ according to the Eq.~(\ref{eq: fedpg local loss});
        \ENDIF
        \STATE Execute inference based on the local GNN to obtain the node embedding and soft labels;\\
        \STATE Perform unlabelled node annotations based on soft labels;\\
        \STATE Obtain topology-aware prototypes $\mathcal{P}^i_{c,h}$ for each class $c$ and hop within RF by Eq.~(\ref{eq: fedpg local prototype generation});\\
        \STATE \uwave{\textit{Upload topology-aware local prototypes}};
    \ENDFOR
    \ENDFOR 
\end{algorithmic}
\end{algorithm}

\begin{algorithm}[h]
\caption{Prototype-guided FGL (FedPG-Server)} 
\label{alg: fedpg_server}
\begin{algorithmic}[1] 
    \FOR {each communication round $t = 1, ..., T$}
        \IF {$t\neq1$}  
        \STATE \uwave{\textit{Receive the uploaded local prototypes}};
        \FOR {each label class $c = 1, ..., |\mathcal{C}|$}
        \FOR {each hop $h = 0, ..., \min(L)$}
        \STATE Obtain positive/negative query sets for each label class and hop within RF;
       \FOR {each server model training epoch $e = 1, ..., E$}
        \STATE Obtain the universal global prototypes through $\tilde{\mathcal{P}}_{c,h} = {GPG}\left(\ddot{\mathcal{P}}_{c,h}, \theta\right)$;
        \STATE Update server-side trainable generation module weights $\theta$ by Eq.~(\ref{eq: fedpg server loss});
    \ENDFOR
        \ENDFOR
        \ENDFOR
        \FOR{each participating client $i=1,\dots,N$ in round $t$}
        \STATE Obtain the personalized global prototype by Eq.~(\ref{eq: fedpg personalized prototype fusion})
        \ENDFOR
        \STATE \uwave{\textit{Broadcast the personalized global prototypes}};
        \ENDIF
    \ENDFOR
\end{algorithmic}
\end{algorithm}

    Table~\ref{tab: algorithm_analysis} provides a theoretical algorithm complexity analysis, where $n$, $m$, $c$, and $f$ are the number of nodes, edges, classes, and feature dimensions, respectively. 
    $s$ is the number of selected augmented nodes and $g$ is the number of generated neighbors.
    $b$ and $T$ are the batch size and training rounds, respectively.
    $k$ and $K$ correspond to the number of times we aggregate features and moments order, respectively. 
    $N$ is the number of participating clients in each training round.
    $t$ represents the number of clients exchanging information with the current client.
    $\omega$ represents the model-wise weight alignment loss term, $Q$ denotes the size of the query set used for CL, $E$ stands for the number of models for ensemble learning, $M$ and $p$ indicate the dimension of the trainable matrix used to mask trainable weights and the prototypes.
    Besides, $P_g$ represents pseudo-graph data stored on the server side.

\begin{table*}[t]
\caption{A summary of recent FGL studies. 
Rank represents the average ranking of prediction accuracy.
}
\scriptsize
\label{tab: review fgl studies}
  \resizebox{\linewidth}{21mm}{
\setlength{\tabcolsep}{1mm}{
\begin{tabular}{cccccc}
\midrule[0.3pt]
Methods            & Model Heterogeneity                                                 & DH (Profile and Topology)                                  & Communication Efficiency                                            &  Subgraph-FL                                               &  Graph-FL                                                  \\ \midrule[0.3pt]
GCFL+~\cite{xie2021gcfl} (NeurIPS 2021)                     & {\XSolidBrush}             & {\XSolidBrush}                         &  (Model Weights)                             & {\XSolidBrush} (Rank 8.0)                    & {\Checkmark} (Rank 3.6)     \\
FedStar~\cite{tan2023fedstar} (AAAI 2023)                   & {\XSolidBrush}             & {\Checkmark}                     &  (Model Weights)                             & {\XSolidBrush} (Rank 6.8)                     & {\Checkmark} (Rank 2.0)    \\
FedSage+~\cite{zhang2021fedsage} (NeurIPS 2021)             & {\XSolidBrush}             & {\XSolidBrush}                         &  (Model Weights + Node Embeddings...)          & {\Checkmark} (Rank 4.8)      & {\XSolidBrush} (Rank 7.6)\\
FGSSL~\cite{huang2023fgssl} (IJCAI 2023)                    & {\XSolidBrush}             & {\Checkmark}                     &  (Model Weights)                             & {\Checkmark} (Rank 6.2)      & {\XSolidBrush} (Rank 4.2)\\
Fed-PUB~\cite{baek2022fedpub} (ICML 2023)                   & {\XSolidBrush}             & {\XSolidBrush}                         &  (Model Weights + Trainable Masks)           & {\Checkmark} (Rank 5.4)      & {\XSolidBrush} (Rank 6.8) \\
FedGTA~\cite{li2023fedgta} (VLDB 2024)                      & {\XSolidBrush}             & {\XSolidBrush}                         &  (Model Weights + Lightweight Statistics)    & {\Checkmark} (Rank 3.2)      & {\XSolidBrush} (Rank 6.4)\\
AdaFGL~\cite{li2024adafgl} (ICDE 2024)                     & {\XSolidBrush}             & {\Checkmark}                     &  (Model Weights)                             & {\Checkmark} (Rank 2.4)      & {\XSolidBrush} (Rank 4.6) \\
FedPG (This Paper)                                          & {\Checkmark}         & {\Checkmark}                     & (Prototypes)                             & {\Checkmark} (Rank 1.0)        & {\Checkmark} (Rank 1.0)     \\ \midrule[0.3pt]
\end{tabular}
}}
\end{table*}

\begin{table*}[t]
\caption{Algorithm complexity analysis for existing prevalent FL and FGL studies. 
}
\label{tab: algorithm_analysis}
\resizebox{\textwidth}{21.5mm}{
\setlength{\tabcolsep}{1.2mm}{

\begin{tabular}{c|ccc|ccc}
\midrule[0.3pt]
Method       & Client Mem.            & Server Mem.            & Inference Mem.      & Client Time.             & Server Time.               & Inference Time           \\ \midrule[0.3pt]
FedAvg       & $O((b+k)f+f^2)$        & $O(Nf^2)$            & $O((b+k)f+f^2)$     & $O(kmf+nf^2)$            & $O(N)$                     & $O(kmf+nf^2)$            \\
FedProx      & $O((b+k)f+\omega f^2)$ & $O(Nf^2)$            & $O((b+k)f+f^2)$     & $O(kmf+nf^2+f^2)$        & $O(N)$                     & $O(kmf+nf^2)$            \\
MOON         & $O((b+k)f+Qf^2)$       & $O(Nf^2)$            & $O((b+k)f+f^2)$     & $O(kmf+nf^2+Qnf)$        & $O(N)$                     & $O(kmf+nf^2)$            \\ \midrule[0.3pt]
GCFL+        & $O((b+k)f+f^2)$        & $O(TNf^2)$      & $O((b+k)f+f^2)$     & $O(kmf+nf^2)$            & $O(N^2(\log(N)+T^2f^2))$ & $O(kmf+nf^2)$            \\
FedStar      & $O(E((b+k)f+f^2))$      & $O(Nf^2)$            & $O(E((b+k)f+f^2))$   & $O(E(kmf+nf^2))$          & $O(N)$                     & $O(E(kmf+nf^2))$          \\
FedSage+     & $O(L((n+sg)f+f^2))$     & $O(LtNf^2)$           & $O(L(n+sg)f+Lf^2)$  & $O(L((m+sg)f+(n+sg)f^2))$ & $O(N)$                     & $O(L((m+sg)f+(n+sg)f^2))$ \\
FGSSL        & $O(Q((b+k)f+f^2))$      & $O(Nf^2)$            & $O(L(b+k)f+Lf^2)$ & $O(Qkmf+Qnf^2)$      & $O(N)$                     & $O(kmf+nf^2)$            \\
Fed-PUB       & $O(M((b+k)f+f^2)+M^2)$      & $O(N(f^2 +M) + P_g)$ & $O(M(b+k)f+Mf^2)$   & $O(Mkmf+Mnf^2)$      & $O(N^2(\log(N)+M^2))$    & $O(Mkmf+Mnf^2)$          \\
FGGP       & $O((n+sg)f+f^2+Qcp)$      & $O(Ncp)$ & $O((b+k)f+f^2)$   & $O((m+sg)f+(n+sg)f^2+Qcp^2)$      & $O(N^2(\log(N)+c^2p^2)+Ncp)$    & $O(kmf+nf^2)$          \\
FedGTA       & $O((b+k)f+f^2+kKc)$    & $O(Nf^2+NkKc)$       & $O((b+k)f+f^2)$     & $O(km(f+knc)+n(f^2+c))$  & $O(N+NkKc)$                & $O(kmf+nf^2)$            \\
AdaFGL       & $O(E((b+k)f+f^2))$      & $O(Nf^2)$            & $O(E((b+k)f+f^2))$   & $O(E(kmf+nf^2))$          & $O(N)$                     & $O(E(kmf+nf^2))$           \\ \midrule[0.3pt]
FedPG (ours) & $O((b+k)f+f^2 + kcp)$  & $O(NQkcp+p^2)$           & $O((b+k)f+f^2)$     & $O(kmf+nf^2+kp)$         & $O(Qkcp^2)$             & $O(kmf+nf^2)$            \\ \midrule[0.3pt]
\end{tabular}
}}

\end{table*}

    For model-agnostic strategies, we choose SGC~\cite{wu2019sgc} as the local model ($k$-step feature propagation), otherwise, we adopt the model architecture ($L$-layer) used in their original paper.
    For the $k$-layer SGC model with batch size $b$, the $\mathbf{X}^{(k)}$ is the $k$-layers propagated feature bounded by a space complexity of $O((b+k)f)$. 
    The overhead for linear regression by multiplying $\mathbf{W}$ is $O(f^2)$.
    In the training stage, the above procedure is repeated to iteratively update the model weights. 
    For the server performing FedAvg, it needs to receive the model weights and the number of samples participating in this round.
    Its space complexity and time complexity are bounded as $O(Nf^2)$ and $O(N)$.
    As discovered by previous studies~\cite{chen2020gbp}, the dominating term is $O(kmf)$ or $O(Lmf)$ when the graph is large since feature learning can be accelerated by parallel computation.

    To improve federated training, local updates are critical.
    For instance, FedProx introduces model weight alignment loss ($O(\omega f^2)$ time complexity). 
    For the CL in MOON, FGSSL, and FGGP, the additional computational cost depends on the size and semantics of the query set, resulting in complexities of $O(Qf^2)$, $Q((b+k)f+f^2)$, $O(Qcp^2)$ respectively.
    This will lead to unacceptable costs as the scale of local data increases. 
    As for ensemble-based FedStar and AdaFGL, which maintain multiple models locally to extract private data semantics and can be bounded by $O(E((b+k)f+f^2))$.
    Furthermore, some methods rely on additional message sharing.
    Specifically, FedSage+ involves client-client message sharing for local subgraph augmentation, leading to a complexity of $O(L((n+sg)f+f^2))$. 
    Fed-PUB maintains a global pseudo-graph on the server side and utilizes locally uploaded weights for personalization, introducing a complexity of $O(N(f^2+M)+P_g)$. 
    In contrast, FedGTA is a lightweight method that utilizes topology-aware soft labels to encode local data, enabling personalized model aggregation on the server.
    However, the additional encoding signal results in a complexity of $O(kKC)$.

    Compared to the above methods, our proposed FedPG exhibits significant scalability advantages in terms of algorithm time-space complexity. 
    Specifically, in addition to the local graph learning common to all methods, FedPG discards a large number of trainable neurons in model weights, reducing complexity from $O(f)$ to $O(p)$ based on semantic encoding extracted from embedding space, where exists $p\ll f$. 
    Meanwhile, in most cases, both $k$ and $c$ are relatively small real values, resulting in negligible computational overhead.
    Despite involving trainable generation of universal global prototypes on the server side, our approach maintains user-friendly computational overhead, owing to lightweight neural architectures and highly compressed prototype representations in low-dimensional space.
    Additionally, the experimental results regarding running efficiency in Sec.~\ref{sec: Performance Comparison} further substantiate our claims from a practical application standpoint.

\begin{table*}[htbp]
\setlength{\abovecaptionskip}{0.2cm}
\caption{The statistical information of the Subgraph-FL experimental datasets.
}
\footnotesize 
\label{tab: datasets_subgraph}
\resizebox{\linewidth}{27mm}{
\setlength{\tabcolsep}{1.8mm}{
\begin{tabular}{cccccccc}
\midrule[0.3pt]
Subgraph-FL & \#Nodes   & \#Edges     & \#Features & \#Classes & \#Train/Val/Test & \#Task               & Description           \\ \midrule[0.3pt]
Cora                   & 2,708     & 5,429       & 1,433      & 7         & 20\%/40\%/40\%   & Transductive      & citation network      \\
CiteSeer               & 3,327     & 4,732       & 3,703      & 6         & 20\%/40\%/40\%   & Transductive      & citation network      \\
PubMed                 & 19,717    & 44,338      & 500        & 3         & 20\%/40\%/40\%   & Transductive      & citation network      \\ \midrule[0.3pt]
Computer               & 13,381    & 245,778     & 767        & 10        & 20\%/40\%/40\%   & Transductive      & co-purchase graph     \\
Physics                & 34,493    & 247,962     & 8,415      & 5         & 20\%/40\%/40\%   & Transductive      & co-authorship graph   \\ \midrule[0.3pt]
ogbn-arxiv             & 169,343   & 2,315,598   & 128        & 40        & 60\%/20\%/20\%   & Transductive      & citation network      \\
ogbn-products          & 2,449,029 & 61,859,140  & 100        & 47        & 10\%/5\%/85\%    & Transductive      & co-purchase graph     \\ 
ogbn-papers100M        & 111,059,956  & 1,615,685,872   & 128 & 172         & 1200k/200k/146k    & Transductive      & citation network \\\midrule[0.3pt]
Flickr                 & 89,250    & 899,756     & 500        & 7         & 44k/22k/22k      & Inductive         & image network         \\
Reddit                 & 232,965   & 11,606,919  & 602        & 41        & 155k/23k/54k     & Inductive         & social network        \\ \midrule[0.3pt]
\end{tabular}
}}
\vspace{0.2cm}
\end{table*}

\begin{table*}[htbp]
    \setlength{\abovecaptionskip}{0.2cm}
    \setlength{\belowcaptionskip}{0.0cm}
\caption{The statistical information of the Graph-FL experimental datasets.
}
\footnotesize 
\label{tab: datasets_graph}
  \resizebox{\linewidth}{13mm}{
\setlength{\tabcolsep}{2.4mm}{
\begin{tabular}{cccccccc}
 \midrule[0.3pt]
Graph-FL    & \#Graphs  & \#Avg.Edges & \#Features & \#Classes & \#Train/Test     & \#Task               & Description           \\ \midrule[0.3pt]
NCI1                   & 4,110     & 32.30       & 37         & 2         & 90\%/10\%        & Graph Classification & molecules network     \\
PROTEINS               & 1,113     & 72.82       & 4          & 2         & 90\%/10\%        & Graph Classification & protein network       \\
IMDB-BINARY            & 1,000     & 96.53       & degree     & 2         & 90\%/10\%        & Graph Classification & movie network         \\
COLLAB                 & 5,000     & 2457.78     & degree     & 3         & 90\%/10\%        & Graph Classification & collaboration network \\ \midrule[0.3pt]
\end{tabular}
}}
\end{table*}

\section{Experiments}
\label{sec: experiments}
    In this section, we present a comprehensive evaluation and aim to answer the following questions:
    \textbf{Q1}: In the multi-level FGL heterogeneity, can FedPG outperform existing SOTA baselines? (Sec.~\ref{sec: Performance Comparison})
    \textbf{Q2}: What is the generalization ability of FedPG in Graph-FL (Sec.~\ref{appendix: Experiments for Graph-FL})? 
    \textbf{Q3}: If FedPG is effective, what contributes to its performance (Sec.~\ref{sec: Ablation Study and In-depth Analysis})?
    \textbf{Q4}: How does the robustness of FedPG when employing noise-based privacy preservation (Sec.~\ref{sec: Robustness in Noise-based Privacy Preservation})?
    \textbf{Q5}: How does FedPG perform under sparse settings and low client participation (Sec.~\ref{sec: Performance under Sparse Settings})?

\subsection{Experimental Setup}
\textbf{Datasets.}
    We evaluate FedPG under Graph- and Subgraph-FL with 14 datasets and 3 distributed simulation strategies.
    To provide a clear illustration, we use Subgraph-FL for a detailed description, including ablation studies, robustness, and efficiency evaluations.
    More details about datasets can be found in Table~\ref{tab: datasets_subgraph}-\ref{tab: datasets_graph}.
    Based on this, we follow previous studies~\cite{zhang2021fedsage,li2023fedgta,baek2022fedpub,li2024adafgl,zhu2024fedtad,zhu2021Non-iid_fl_survey1, matsuda2022Non-iid_fl_survey2, li2022Non-iid_fl_survey3} to provide Louvain-~\cite{blondel2008louvain} and Metis-based~\cite{karypis1998metis} distributed simulation for subgraph-FL, and we employ label Dirichlet and uniform distribution for graph-FL.

\begin{table*}[t]
\caption{Subgraph-FL test accuracy (\%).
The best result is \textbf{bold}.
The second result is \underline{underlined}.}
\label{tab: subgraph-fl end-to-end comparison}
\resizebox{\linewidth}{37mm}{
\setlength{\tabcolsep}{2mm}{
\begin{tabular}{ccccccccc|cc}
\midrule[0.3pt]
Method   & Cora     & CiteSeer & PubMed   &  \begin{tabular}[c]{@{}c@{}}Amazon\\ Computer\end{tabular} &  \begin{tabular}[c]{@{}c@{}}Coauthor\\ Physics\end{tabular}  & \begin{tabular}[c]{@{}c@{}}ogbn\\ arxiv\end{tabular}    & \begin{tabular}[c]{@{}c@{}}ogbn\\ products\end{tabular} & \begin{tabular}[c]{@{}c@{}}ogbn\\ papers100M\end{tabular} & Flickr   & Reddit   \\ \midrule[0.3pt]
FedAvg   & 82.8±0.3 & 70.4±0.2 & 86.2±0.1 & 80.6±0.4 & 89.6±0.6 & 71.4±0.5 & 76.7±0.3 & 62.4±0.2& 48.6±0.3 & 92.9±0.1 \\
FedProx   & 83.0±0.4 & 70.2±0.2 & 86.5±0.2 & 80.8±0.4 & 89.5±0.6 & 71.6±0.6 & 76.5±0.4 & 61.8±0.3& 48.9±0.2 & 93.0±0.0 \\
MOON     & 82.4±0.4 & 70.5±0.3 & 85.8±0.2 & 80.9±0.5 & 89.1±0.7 & 71.8±0.6 & 76.3±0.5 & 63.2±0.4& 49.3±0.4 & 92.8±0.1 \\ \midrule[0.3pt]
FedProto & 76.7±0.3 & 65.2±0.2 & 81.7±0.1 & 74.5±0.3 & 83.2±0.4 & 62.6±0.4 & 67.8±0.3 & 56.8±0.3& 42.9±0.3 & 87.6±0.0 \\
FPL     & 76.5±0.4& 65.7±0.3 & 82.0±0.1 & 74.8±0.4 & 82.8±0.6 & 63.3±0.5 & 67.6±0.3 & 58.5±0.2& 43.0±0.5 & 87.9±0.1 \\
FedNH    & 77.2±0.4 & 65.4±0.2 & 81.4±0.1 & 75.4±0.3 & 83.0±0.5 & OOT & OOT & OOM& 42.4±0.4 & OOT \\
FedTGP   & 76.5±0.5 & 65.5±0.3 & 81.9±0.2 & 74.9±0.4 & 82.6±0.7 & 64.8±0.6 & 72.4±0.4 & 60.3±0.5& 42.8±0.6 & 88.5±0.1 \\ \midrule[0.3pt]
FedSage+ & 83.9±0.5 & 71.4±0.4 & 87.0±0.3 & 81.2±0.5 & 90.0±0.6 & 72.7±0.7 & OOM      & OOM& 50.2±0.6 & OOM      \\
FGSSL    & 83.6±0.4 & 70.9±0.3 & 86.6±0.2 & 81.4±0.4 & 89.5±0.7 & 72.0±0.6 & OOM      & OOM& 49.8±0.5 & OOM      \\
Fed-PUB  & 83.5±0.4 & 71.1±0.3 & 86.9±0.2 & 81.5±0.4 & \underline{90.3±0.5} & 72.3±0.5 & 77.6±0.5 & OOM& 50.4±0.4 & 93.6±0.1 \\
FedGTA   & 84.0±0.3 & \underline{71.6±0.2} & 86.8±0.1 & \underline{81.8±0.3} & 90.1±0.4 & 72.5±0.4 & \underline{79.2±0.4} & 65.7±0.3& \underline{50.8±0.4} & {93.9±0.0} \\
AdaFGL   & \underline{84.2±0.4} & 71.5±0.4 & \underline{87.1±0.2} & 81.5±0.5 & 89.9±0.5 & \underline{72.9±0.6} & 79.0±0.6 & OOM& {50.7±0.5} & \underline{94.1±0.1} \\ \midrule[0.3pt]
FGGP     & 83.1±0.4 & 70.8±0.3 & 86.5±0.2 & 80.4±0.5 & 88.9±0.5 & 71.9±0.5 & OOM & OOM& 47.9±0.5 & OOM \\
FedPG    & \textbf{86.5±0.4} & \textbf{73.6±0.3} & \textbf{89.4±0.2} & \textbf{83.9±0.3} & \textbf{92.2±0.4} & \textbf{74.1±0.5} & \textbf{81.7±0.4} & \textbf{67.4±0.4}& \textbf{52.3±0.4} & \textbf{95.6±0.0} \\ 
+ (10\% Noise)    & \cellcolor{gray!25}{86.1±0.5} & \cellcolor{gray!25}{73.3±0.3} & \cellcolor{gray!25}{89.2±0.3} & \cellcolor{gray!25}{83.6±0.3} & \cellcolor{gray!25}{92.0±0.4} & \cellcolor{gray!25}{74.0±0.4} & \cellcolor{gray!25}{81.5±0.3} & \cellcolor{gray!25}{67.1±0.3}& \cellcolor{gray!25}{52.2±0.3} & \cellcolor{gray!25}{95.5±0.1} \\ \midrule[0.3pt]
\end{tabular}
}}
\end{table*}

\noindent
\textbf{Baselines.}
    For FL, we compare FedPG with FedAvg~\cite{mcmahan2017fedavg}, FedProx~\cite{li2020fedprox}, Scaffold~\cite{karimireddy2020scaffold}, MOON~\cite{li2021moon}, and FedDC~\cite{gao2022feddc}.
    To evaluate MH, FedDistill~\cite{jeong2018feddistill}, FedGen~\cite{zhu2021fedgen}, and FedKD~\cite{wu2022fedkd} are taken into consideration.
    Regarding FPL, we include FedProto~\cite{tan2022fedproto}, FedNH~\cite{dai2023fednh}, FPL~\cite{huang2023fpl}, and FedTGP~\cite{zhang2024fedtgp}.
    As for FGL, compare FedPG with GCFL+~\cite{xie2021gcfl}, FedStar~\cite{tan2023fedstar}, FedSage+~\cite{zhang2021fedsage}, FGSSL~\cite{huang2023fgssl}, Fed-PUB~\cite{baek2022fedpub}, FedGTA~\cite{li2023fedgta}, AdaFGL~\cite{li2024adafgl}, and FGGP~\cite{wan2024fggp}.
    To comprehensively reveal the effectiveness of FedPG while avoiding complex charts, we experiment with multiple baselines in separate modules.  

    To alleviate the randomness and ensure a fair comparison, we repeat each experiment 10 times for unbiased performance.
    Unless otherwise stated, for Graph- and Subgraph-FL, we respectively adopt GIN and GAMLP as the backbone and employ the uniform- and Metis-based 10-client simulation.
    For the extension of FedPG to the Graph-FL, please refer to Sec.~\ref{appendix: Experiments for Graph-FL}.
    Meanwhile, to evaluate the effectiveness of our proposed FedPG for MH (detailed in Sec.~\ref{sec: introduction}), we consider Subgraph/Graph-FL with 5x client partitions.
    Then, we randomly and equally allocate the following 5 local backbones: GCN~\cite{kipf2016gcn}, GIN~\cite{xu2018gin}, GraphSAGE~\cite{hamilton2017graphsage}, SGC~\cite{wu2019sgc}, and GCNII~\cite{chen2020gcnii}.

\subsection{Experiment Environment}
\label{appendix: Experiment Environment}
    The machine with Intel(R) Xeon(R) Gold 6240 CPU @ 2.60GHz, and NVIDIA A100 80GB PCIe and CUDA 12.2. 
    The operating system is Ubuntu 18.04.6.
    As for software versions, we use Python 3.8.10, Pytorch 1.10.0, CUDA 11.7.0. 

\subsection{Hyperparameter Settings}
\label{appendix: Hyperparameter Settings}
    The hyperparameters in the baselines are set according to the original paper if available.
    Otherwise, we perform a hyperparameter search via the Optuna~\cite{akiba2019optuna} auto machine learning library.
    For our proposed FedPG, we explore $\epsilon\in [0,1]$ and $\vartriangle\!s\in [0,1]$ for the adaptive maximum discrepancy and hop-based sampling, where $\vartriangle\!s$ is the ratio of additional sampling from other hops based on the number of current query samples.
    The personalized fusion factor $\alpha$, similarity threshold $\lambda$, and loss factor $\mu$ are explored within the ranges of 0 to 1.

\subsection{Subgraph-FL Comparison}
\label{sec: Performance Comparison}
\vspace{0.05cm}

\noindent
\textbf{End-to-end Comparison in DH.}
    We report the experimental results in Table~\ref{tab: subgraph-fl end-to-end comparison} and observe that FedPG consistently outperforms other baselines, including its ability to predict unseen nodes, as highlighted by the inductive performance.
    Specifically, FedPG achieves improvements of 2.76\% and 3.85\% over the competitive Fed-PUB and AdaFGL. 
    While FGGP pioneer prototype-based FGL, it neglects topology-driven optimal prototypes, resulting in its weak performance.
    In terms of scalability, FedSage+, FGSSL, and FGGP struggle in large-scale scenarios due to their prohibitive complexity in local graph augmentation, often resulting in out-of-memory (OOM) errors.
    FedNH aims to maximize inter-class distances for prototype initialization, but the interior point method used to solve this optimization problem incurs significant computational costs, especially on datasets with a large number of label classes.
    This often results in exceeding two hours without completion, denoted as out-of-time (OOT).
    Moreover, CV-based FPL methods distort prototype semantics and hinder optimization due to the neglect of topology, resulting in weak performance.
    This observation supports our claims in Sec.~\ref{sec: introduction}.

\vspace{0.02cm}
\noindent
\textbf{Heterogeneous Collaboration in MH.}
    To the best of our knowledge, most FGL methods fail to handle MH due to the direct aggregation of uploaded models on the server. 
    Therefore, we compare with FGGP and baselines that implement MH in the CV.
    According to the results in Table~\ref{tab: subgraph-fl mh}, we observe that the negative impact of ignoring graph topology by CV-based FPL methods, as presented in Table~\ref{tab: subgraph-fl end-to-end comparison}, is further amplified. 
    This is attributed to the exacerbation of the optimization dilemma in graph semantic perception by MH. 
    Additionally, compared to the most competitive FGGP in the MH scenario, FedPG achieves an average performance improvement of 12.4\%. 
    Our achievement can be attributed to obtaining optimal graph prototypes within the RF guidance. 
    This approach effectively measures nuanced semantic variances among clients, encompassing model architectures and data distributions, thereby facilitating efficient federated training. 

\begin{table}[t]  
    \setlength{\abovecaptionskip}{0.2cm}
    \setlength{\belowcaptionskip}{-0.2cm}
  \centering
  \caption{Test accuracy (\%) in MH with Louvain split.}
  \label{tab: subgraph-fl mh}
  \resizebox{\linewidth}{20mm}{
\setlength{\tabcolsep}{2.8mm}{
\begin{tabular}{c|cc|cc}
\midrule[0.3pt]
Datasets($\rightarrow$) & \multicolumn{2}{c|}{ogbn-arxiv} & \multicolumn{2}{c}{Flickr} \\ \midrule[0.3pt]
Method($\downarrow$)    & 10 Client    & 20 Client   & 10 Client    & 20 Client   \\ \midrule[0.3pt]
FedKD                   & 51.2±0.8     & 49.1±0.7    & 40.5±0.5     & 39.2±0.6    \\
FedGen                  & 48.1±0.6     & 45.5±0.8    & 37.8±0.6     & 37.0±0.5    \\
FedDistill              & 50.3±0.7     & 47.0±0.5    & 39.4±0.7     & 38.6±0.8    \\
FedProto                & 58.9±0.6     & 55.7±0.8    & 41.2±0.5     & 40.9±0.6    \\
FedTGP                  & 60.3±0.7     & 57.8±0.8    & 42.3±0.6     & 41.5±0.6    \\
FGGP                    & 65.5±0.8     & 63.3±0.7    & 44.9±0.6     & 42.4±0.7    \\
FedPG (Ours)                   & \textbf{71.2±0.6}     & \textbf{70.5±0.5}    & \textbf{51.6±0.5}     & \textbf{50.8±0.6}    \\ \midrule[0.3pt]
\end{tabular}
}}
\end{table}

\begin{table}[t]  
    \setlength{\abovecaptionskip}{0.2cm}
    \setlength{\belowcaptionskip}{-0.cm}
  \centering
  \caption{Running efficiency on Physics with Louvain split.}
  \label{tab: subgraph-fl efficiency report}
  \resizebox{\linewidth}{21mm}{
\setlength{\tabcolsep}{2.5mm}{
\begin{tabular}{ccccc}
\midrule[0.3pt]
Method   & Acc(\%)  & Parameters & Time     & Communication  \\ \midrule[0.3pt]
Scaffold & 89.2±0.4 & 538k   & 149.78s  & 1076k \\
FedDC    & 89.3±0.6 & 538k   & 178.42s  & 1346k \\
FedTGP   & 86.5±0.4 & 539k   & 75.39s   & 0.38k \\\midrule[0.3pt]
FedSage+ & 91.0±0.5 & 1296k  & 1493.55s & 1784k \\
Fed-PUB  & 90.7±0.4 & 1076k  & 384.64s  & 1076k \\
FedGTA   & 90.9±0.3 & 538k   & 126.57s  & 539k  \\
AdaFGL   & 90.6±0.5 & 964k   & 167.83s  & 538k  \\ \midrule[0.3pt]
FGGP     & 89.4±0.6 & 538k   & 321.16s  & 0.38k \\ 
FedPG (Ours)    & 92.8±0.3 & 539k   & 98.18s   & 1.14k \\ \midrule[0.3pt]
\end{tabular}
}}
\end{table}

\begin{figure*}[t]
	\centering
    \setlength{\abovecaptionskip}{0.2cm}
    \setlength{\belowcaptionskip}{0.1cm}
  \includegraphics[width=\textwidth]{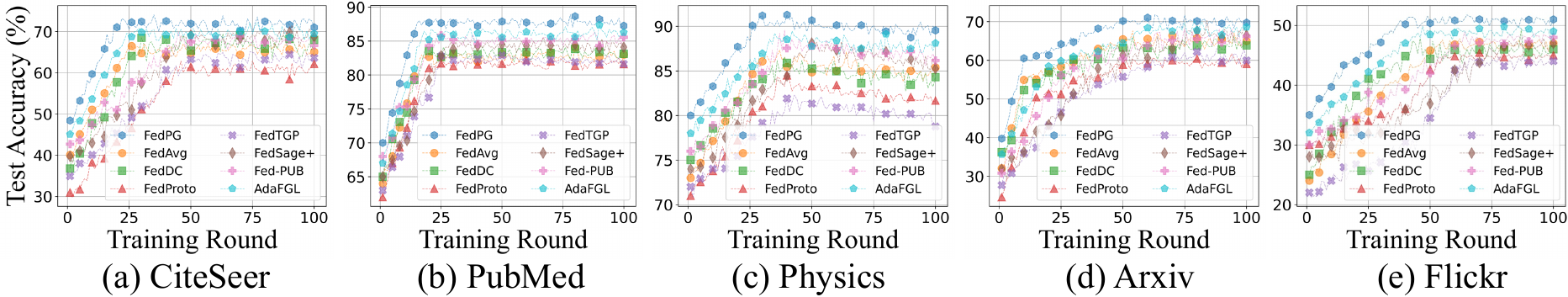}
  \caption{
    Convergence curves with Louvain split and GCNII backbone.
}
  \label{fig: exp_converge}
\end{figure*}

\noindent
\textbf{Efficiency Comparison in CH.}
    Building upon the algorithm complexity analysis shown in Sec.~\ref{appendix: FedPG Algorithm and Complexity Analysis}, we present a practical efficiency report in Table~\ref{tab: subgraph-fl efficiency report}, where shows the trainable parameters at both the client and server sides, the overall runtime, and the communication costs among clients or between clients and the server.
    We observe that although FedTGP and FGGP exhibit the lowest communication, their lack of perception of intricate graph semantic and complex local augmentation results in weak and time-consuming performance. 
    Meanwhile, while FedSage+, FedGTA, and AdaFGL achieve competitive performance, their heavy reliance on message sharing leads to significant communication and computational complexity.
    In contrast, FedPG achieves SOTA performance while enjoying lower communication and higher efficiency.

\noindent
\textbf{Training Efficiency.}
    In FedPG, we aim to derive optimal prototypes by
    (1) Each client adaptively generates multi-granularity representations for prototypes of each class in a topology-aware manner to fully extract local graph contexts. 
    (2) The server customizes global prototypes for each client through topology-guided CL and personalized mechanisms for broadcasting. 
    The core is to fully consider the graph context in FPL to maximize training efficiency, reflected in faster and more stable convergence curves and higher accuracy. 
    
    To validate our claims, we provide comprehensive experimental results in Fig.~\ref{fig: exp_converge}.
    In PubMed, we observe that all methods demonstrate rapid convergence and ensure stable performance. However, on datasets like ogbn-arxiv and Flickr, achieving optimal predictions requires more communication rounds and is accompanied by unstable optimization.
    Although this difficulty varies across different datasets due to differences in data quantity and quality, FedPG consistently demonstrates SOTA performance from start to finish, regardless of dataset intricacies.
    For example, in the Physics dataset, FedPG nearly reaches converged performance by the 25th epoch and maintains stability throughout subsequent training.

\begin{table*}[t]
\caption{Graph-FL test accuracy (\%).
The best result is \textbf{bold}.
The second result is \underline{underlined}.}
\vspace{-0.15cm}
\label{tab: graph-fl end-to-end comparison1}
  \resizebox{\linewidth}{25mm}{
\setlength{\tabcolsep}{3.6mm}{
\begin{tabular}{c|ccccc|ccccc}
\midrule[0.3pt]
Datasets $\rightarrow$   & \multicolumn{5}{c|}{NCI1}                                                                                              & \multicolumn{5}{c}{PROTEINS}                                                                                           \\ \midrule[0.3pt]
Simulation $\rightarrow$ & \multicolumn{3}{c|}{Non-iid (ACC)}                                             & \multicolumn{2}{c|}{iid}              & \multicolumn{3}{c|}{Non-iid (ACC)}                                             & \multicolumn{2}{c}{iid}               \\ \midrule[0.3pt]
Method ($\downarrow$)    & 0.3               & 0.5               & \multicolumn{1}{c|}{1}                 & ACC               & AUC               & 0.3               & 0.5               & \multicolumn{1}{c|}{1}                 & ACC               & AUC               \\ \midrule[0.3pt]
FedAvg                   & 68.7±5.4          & 70.6±4.0          & \multicolumn{1}{c|}{72.8±3.4}          & 74.5±2.2          & 73.8±1.7          & {67.4±4.5} & {69.5±3.9} & \multicolumn{1}{c|}{{72.5±2.8}} & {73.4±2.5} & {73.9±2.6} \\
FedDC                    & 68.5±4.8          & 70.1±4.6          & \multicolumn{1}{c|}{72.4±3.5}          & 73.6±2.8          & 73.5±2.4          & 66.5±5.2          & 69.2±4.0          & \multicolumn{1}{c|}{72.9±3.0}          & 74.6±2.6          & 75.0±2.4          \\ \midrule[0.3pt]
FedProto                 & 67.4±3.8          & 69.2±3.5          & \multicolumn{1}{c|}{72.2±3.2}          & 73.8±1.9          & 74.2±1.7          & {67.1±3.4} & {68.6±2.9} & \multicolumn{1}{c|}{{71.8±2.4}} & {73.1±1.9} & {72.8±2.1} \\
FedTGP                   & 66.0±5.2          & 68.7±4.1          & \multicolumn{1}{c|}{72.5±2.5}          & 73.5±2.0          & 74.0±2.3          & 66.9±4.3          & 68.3±3.6          & \multicolumn{1}{c|}{71.5±2.7}          & 72.8±2.4          & 73.0±2.5          \\ \midrule[0.3pt]
GCFL+                    & \underline{74.3±2.0}          & \underline{75.4±2.3}          & \multicolumn{1}{c|}{76.1±1.8}          & 77.3±1.6          & 77.6±1.8          & \underline{71.5±3.4} & \underline{73.2±3.6} & \multicolumn{1}{c|}{{75.5±2.4}} & 76.8±2.8          & {76.5±3.3} \\
FedStar                  & 71.5±4.9          & 74.2±3.8          & \multicolumn{1}{c|}{\underline{76.8±2.6}}          & \underline{77.9±1.5}          & \underline{78.2±1.3}          & 68.6±4.9          & 71.8±4.2          & \multicolumn{1}{c|}{\underline{75.9±3.0}}          & \underline{77.5±3.2}          & \underline{77.2±3.5}          \\ \midrule[0.3pt]
FGGP                     & 73.6±3.7          & 75.0±2.5          & \multicolumn{1}{c|}{76.5±2.4}          & 77.6±2.1          & 77.1±2.2          & 69.3±4.5          & 70.2±3.9          & \multicolumn{1}{c|}{73.4±3.3}          & 74.3±2.6          & 74.0±2.7          \\
FedPG                    & \textbf{77.4±2.6} & \textbf{78.2±2.3} & \multicolumn{1}{c|}{\textbf{79.3±2.0}} & \textbf{79.8±1.7} & \textbf{79.6±1.6} & \textbf{76.5±3.4} & \textbf{77.4±3.0} & \multicolumn{1}{c|}{\textbf{78.6±2.5}} & \textbf{79.2±2.2} & \textbf{79.4±2.0} \\ \midrule[0.3pt]
\end{tabular}
}}
\end{table*}

\begin{table*}[t]
\caption{Graph-FL test accuracy (\%).
The best result is \textbf{bold}.
The second result is \underline{underlined}.}
\vspace{-0.15cm}
\label{tab: graph-fl end-to-end comparison2}
  \resizebox{\linewidth}{25mm}{
\setlength{\tabcolsep}{3.6mm}{
\begin{tabular}{c|ccccc|ccccc}
\midrule[0.3pt]
Datasets $\rightarrow$   & \multicolumn{5}{c|}{IMDB-BINARY}                                               & \multicolumn{5}{c}{COLLAB}                                                    \\ \midrule[0.3pt]
Simulation $\rightarrow$ & \multicolumn{3}{c|}{Non-iid (ACC)}                  & \multicolumn{2}{c|}{iid} & \multicolumn{3}{c|}{Non-iid (ACC)}                  & \multicolumn{2}{c}{iid} \\ \midrule[0.3pt]
Method ($\downarrow$)    & 0.3      & 0.5      & \multicolumn{1}{c|}{1}        & ACC         & AUC        & 0.3      & 0.5      & \multicolumn{1}{c|}{1}        & ACC        & AUC        \\ \midrule[0.3pt]
FedAvg                   & 72.4±4.2 & 74.8±3.4 & \multicolumn{1}{c|}{77.2±3.0} & 78.6±2.7    & 78.9±2.6   & 74.1±2.9 & 75.6±2.4 & \multicolumn{1}{c|}{77.2±1.8} & 77.8±1.6   & 78.0±1.9   \\
FedDC                    & 72.2±5.9 & 74.5±5.2 & \multicolumn{1}{c|}{76.8±4.2} & 77.9±3.8    & 78.3±3.5   & 73.4±4.6 & 75.2±3.5 & \multicolumn{1}{c|}{77.0±2.3} & 77.5±2.4   & 77.9±2.2   \\ \midrule[0.3pt]
FedProto                 & 71.9±3.5 & 73.2±3.1 & \multicolumn{1}{c|}{75.6±2.4} & 76.3±2.2    & 76.7±2.3   & 72.2±2.6 & 73.0±1.9 & \multicolumn{1}{c|}{74.2±1.5} & 74.5±1.3   & 74.2±1.5   \\
FedTGP                   & 72.4±3.8 & 73.0±3.3 & \multicolumn{1}{c|}{75.2±3.0} & 76.6±2.4    & 76.5±2.8   & 71.7±2.8 & 72.8±2.2 & \multicolumn{1}{c|}{75.0±1.4} & 75.3±1.6   & 75.5±1.9   \\ \midrule[0.3pt]
GCFL+                    & \underline{76.2±5.4} & \underline{77.9±4.2} & \multicolumn{1}{c|}{79.5±3.7} & 80.4±3.5    & 79.8±3.9   & 73.5±3.5 & 74.5±2.7 & \multicolumn{1}{c|}{77.6±1.8} & 78.2±1.5   & 78.0±1.8   \\
FedStar                  & 75.0±6.3 & 77.2±5.6 & \multicolumn{1}{c|}{\underline{80.3±4.5}} & \underline{81.2±3.9}    & \underline{80.9±3.8}   & \underline{73.9±4.9} & \underline{75.0±3.8} & \multicolumn{1}{c|}{\underline{78.5±3.0}} & \underline{79.4±2.8}   & \underline{79.9±2.7}   \\ \midrule[0.3pt]
FGGP                     & 71.6±4.3 & 73.0±3.8 & \multicolumn{1}{c|}{76.9±2.5} & 77.4±2.4    & 77.5±2.5   & 72.9±3.6 & 74.7±3.1 & \multicolumn{1}{c|}{76.1±2.2} & 76.5±1.9   & 76.9±2.1   \\
FedPG                    & \textbf{78.0±3.7} & \textbf{79.2±3.2} & \multicolumn{1}{c|}{\textbf{80.9±3.0}} & \textbf{81.5±2.6}    & \textbf{82.2±2.9}   & \textbf{78.0±2.8} & \textbf{78.9±2.2} & \multicolumn{1}{c|}{\textbf{80.0±1.6}} & \textbf{80.4±1.5}   & \textbf{80.1±1.6}   \\ \midrule[0.3pt]
\end{tabular}
}}
\vspace{0.2cm}
\end{table*}

\subsection{Graph-FL Comparison.}
\label{appendix: Experiments for Graph-FL}

\noindent
    \textbf{Extend FedPG to Graph-FL.}
    Compared to Subgraph-FL with structured relationships, defining effective neighborhoods is challenging in Graph-FL.
    In our implementation, we utilize appropriate representation distance metric functions and decaying thresholds to identify potentially similar independent graphs as a generalized neighborhood, which uncovers multi-hop neighborhoods, beyond 1-hop connections. 
    This strategy is inspired by the complex network research~\cite{mcpherson200homophily_theory1, M2003Mixing_homophily_theory2, 0Networks_homophily_theory3}.
    Specifically, we posit that independently represented graphs sharing similarities should exhibit underlying correlations. 
    This belief arises from the effective representations of GNNs in graph classification, offering insights into structural nuances and node feature interactions.
    By emphasizing the identification and utilization of these similarities, our approach aims to improve neighborhood definition effectiveness in Graph-FL scenarios.
    To this end, in Graph-FL, the natural extension of FedPG can be achieved by adding a generalized neighborhood construction step after obtaining the encoded representations of each independent graph locally.

\noindent
\textbf{End-to-end Comparison in Graph-FL.}
    We present comprehensive experimental results in Table~\ref{tab: graph-fl end-to-end comparison1} and Table~\ref{tab: graph-fl end-to-end comparison2} under various data simulation settings. 
    In our implementation, 0.3, 0.5, and 1 represent 0.3-, 0.5-, and 1-Dirichlet Non-iid settings. 
    Conversely, iid denotes a distributed graph storage simulation achieved by uniformly sampling label distributions. 
    Based on this, we employ ACC and AUC for comprehensive assessment.
    Experimental results demonstrate that FedPG significantly outperforms SOTA baselines across all scenarios. 
    Compared to other FPL methods, FedPG's explicit consideration of topology-driven optimal prototype generation positively impacts federated collaborative optimization. 
    Additionally, our in-depth analysis yields the following insights: 
    
    (1) Non-iid data settings pose collaborative optimization challenges for all methods. 
    However, FPL mitigate this issue to some extent by directly extracting label class knowledge without aggregating model weights.
    (2) For FPL, the intrinsic semantic content of sample features is crucial, directly affecting the accuracy of extracting global semantic consensus to guide local updates. 
    This aspect is confirmed in experimental results: FPL methods perform significantly better on NCI1 and PROTEINS with genuine semantic content compared to IMBD-BINARY and COLLAB where node degree is used as a feature. 
    While local graph learning models based on node degree can encode structural insights to some extent within embeddings, they still lack genuine semantic knowledge compared to other methods.
    Despite challenges posed by the absence of genuine semantics, our proposed FedPG still stands out and achieves satisfactory predictive performance.

\subsection{Ablation Study and In-depth Analysis}
\label{sec: Ablation Study and In-depth Analysis}

\begin{table}[t]  
    \setlength{\abovecaptionskip}{0.2cm}
    \setlength{\belowcaptionskip}{-0.2cm}
  \centering
  \caption{Ablation study performance (\%).}
  \label{tab: subgraph-fl ablation study}
    \resizebox{\linewidth}{24mm}{
\setlength{\tabcolsep}{2.2mm}{
\begin{tabular}{cccc}
\midrule[0.3pt]
Module           & Cora     &  ogbn-products   & Flickr\\ \midrule[0.3pt]
w/o RF Context  & 84.4±0.5 & 79.5±0.3 & 50.8±0.4 \\
w/o Ada. Margin   & 85.0±0.3 & 80.0±0.4 & 51.5±0.5 \\
w/o Personalized & 84.2±0.6 & 79.3±0.4 & 50.7±0.3 \\
FedPG            & 86.5±0.4 & 81.7±0.2 & 52.3±0.4 \\ \midrule[0.3pt]
\midrule[0.3pt]
Module           & Computer     &  ogbn-papers100M   & Reddit\\ \midrule[0.3pt]
w/o RF Context  & 81.6±0.2 & 66.2±0.3 & 93.9±0.1 \\
w/o Ada. Margin   & 82.4±0.5 & 66.5±0.5 & 94.0±0.0 \\
w/o Personalized & 82.1±0.4 & 66.0±0.3 & 93.7±0.1 \\
FedPG            & 83.9±0.3 & 67.4±0.4 & 95.6±0.0 \\ \midrule[0.3pt]
\end{tabular}
}}
\end{table}

\noindent
    \textbf{Prototype Learning}.
    We investigate the contributions of RF-prompted context (RF Context) and Margin-based Enhancement (Ada. Margin) in Eq.(\ref{eq: fedpg server loss}).
    Specifically, RF Context enhances the generalization of global prototypes by providing rich graph context samples for subsequent CL, which brings 2.21\% average performance improvement.
    Regarding Ada. Margin, it enhances the inter-class semantic boundaries while maintaining intra-class semantic consistency by topology-guided CL. 
    These modules improve the performance on ogbn-products from 80 to 81.7 while reducing the variance from 0.4 to 0.2.
    Based on this, we further investigate RF Context and Ada. Margin from hyperparameter analysis perspective
    
\begin{figure}[t]
\centering
    \setlength{\abovecaptionskip}{0.25cm}
    \setlength{\belowcaptionskip}{-0.2cm}
  \includegraphics[width=\linewidth,scale=1.00]{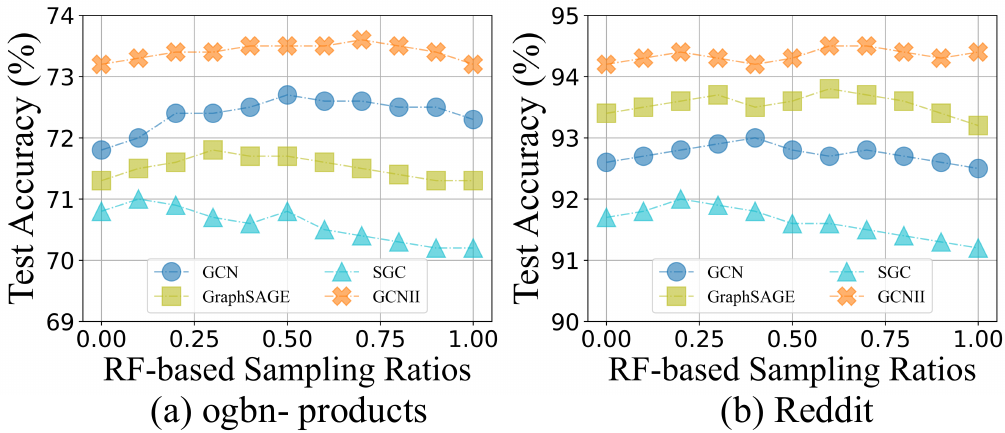}
  \caption{
    Performance with RF prompts.
}
  \label{fig: exp_sampling}
\end{figure}

    As highlighted in Sec.~\ref{sec: Adaptive Global Prototype Generation}, RF-prompted Context samples prototypes of the same class from different hops to supplement the query set.
    In Fig.~\ref{fig: exp_sampling}, We visualize the effect of sampling ratio on training.
    We observe that both insufficient and excessive sampling adversely affect the final performance and the quality of local prototypes fundamentally depends on the local backbone's reasoning ability.
    Notably, GraphSAGE is specifically designed for inductive scenarios, hence exhibiting slightly better performance on Reddit compared to GCN, while slightly underperforming on the ogbn-products.

\begin{figure}[t]  
\centering
    \setlength{\abovecaptionskip}{0.2cm}
    \setlength{\belowcaptionskip}{-0.2cm}
  \includegraphics[width=\linewidth,scale=1.00]{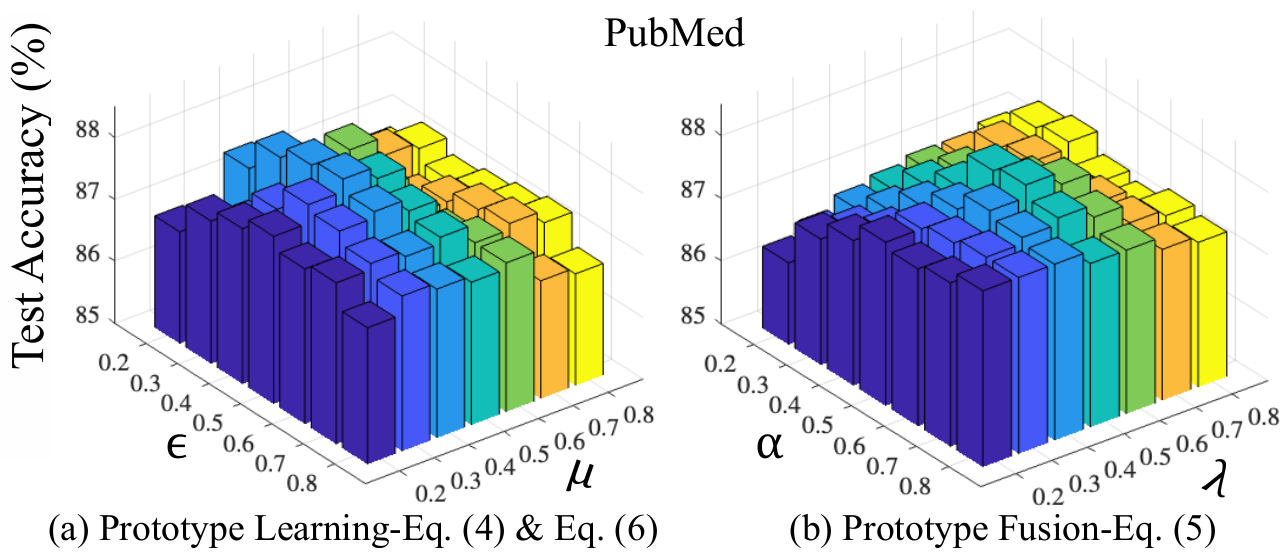}
  \caption{
    Sensitive analysis with Louvain split and GCN backbone.
}
  \label{fig: exp_hyperparameter}
\end{figure}

    Moreover, in Fig.~\ref{fig: exp_hyperparameter} (a), we further investigate Ada. Margin and prototype alignment loss through $\epsilon$ and $\mu$.
    Experimental results indicate that excessively small or large $\epsilon$ hinders Ada. Margin in discerning prototypes.
    Notably, balancing loss is essential for local updates while acquiring global prototype knowledge. 
    Therefore, we most recommend prioritizing $\epsilon=\mu=0.5$, followed by minor manual adjustments.

\noindent
     \textbf{Prototype Fusion}.
    As mentioned in Sec.~\ref{sec: Adaptive Global Prototype Generation}, the customization of global prototypes maximizes local training efficiency, resulting in an average performance boost of 2.48\% across six datasets.
    To further validate the effectiveness of this module, we delve into two crucial parameters of personalized prototype fusion in Fig. \ref{fig: exp_hyperparameter} (b). 
    Experimental results suggest that optimal performance often requires moderate $\alpha$ and larger $\lambda$. 
    Specifically, the former $\alpha$ ensures a balance between personalized and global prototypes during fusion, while the latter $\lambda$ ensures fusion occurs only among semantically similar local clients to prevent knowledge confusion.

\begin{figure}[t]  
\centering
    \setlength{\abovecaptionskip}{0.2cm}
    \setlength{\belowcaptionskip}{-0.3cm}
  \includegraphics[width=\linewidth,scale=1.00]{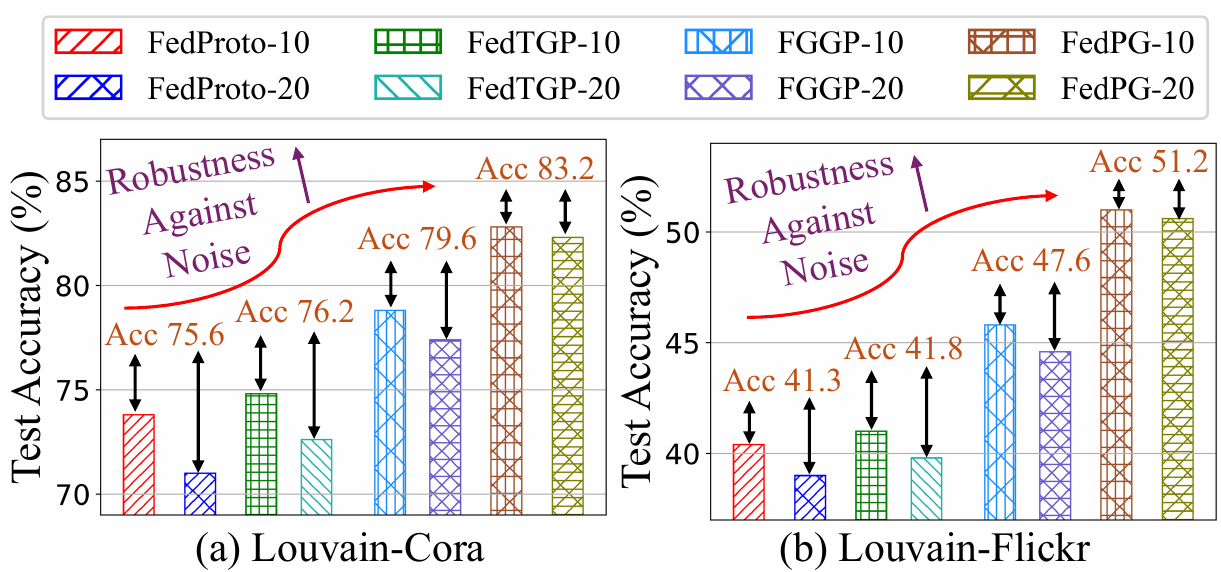}
  \caption{
    Noise-based privacy preservation with GAT backbone.
}
  \label{fig: exp_noise}
\end{figure}

\subsection{Noise-based Privacy Preservation}
\label{sec: Robustness in Noise-based Privacy Preservation}
    Despite the distinction between prototypes and raw profiles from the statistical perspective, directly exchanging them may still raise privacy concerns. 
    Therefore, we add noise to prototypes shown in Table~\ref{tab: subgraph-fl end-to-end comparison} and Fig.~\ref{fig: exp_noise}, where 10 and 20 respectively denote randomly selecting 10\% and 20\% of prototype dimensions for noise addition.
    Experimental results show that FedPG exhibits natural robustness.
    This is attributed to
    (1) Client-side topology-aware encoding provides multi-granularity prototypes, allowing both clients and servers to adaptively obtain positive signals from different sources and resist noise interference. 
    (2) Server-side prototype generation adaptively filters noisy prototypes through CL and personalized mechanisms to ensure high-quality global prototypes.

\begin{figure}[t]
\centering
    \setlength{\abovecaptionskip}{0.2cm}
    \setlength{\belowcaptionskip}{-0.3cm}
  \includegraphics[width=\linewidth,scale=1.00]{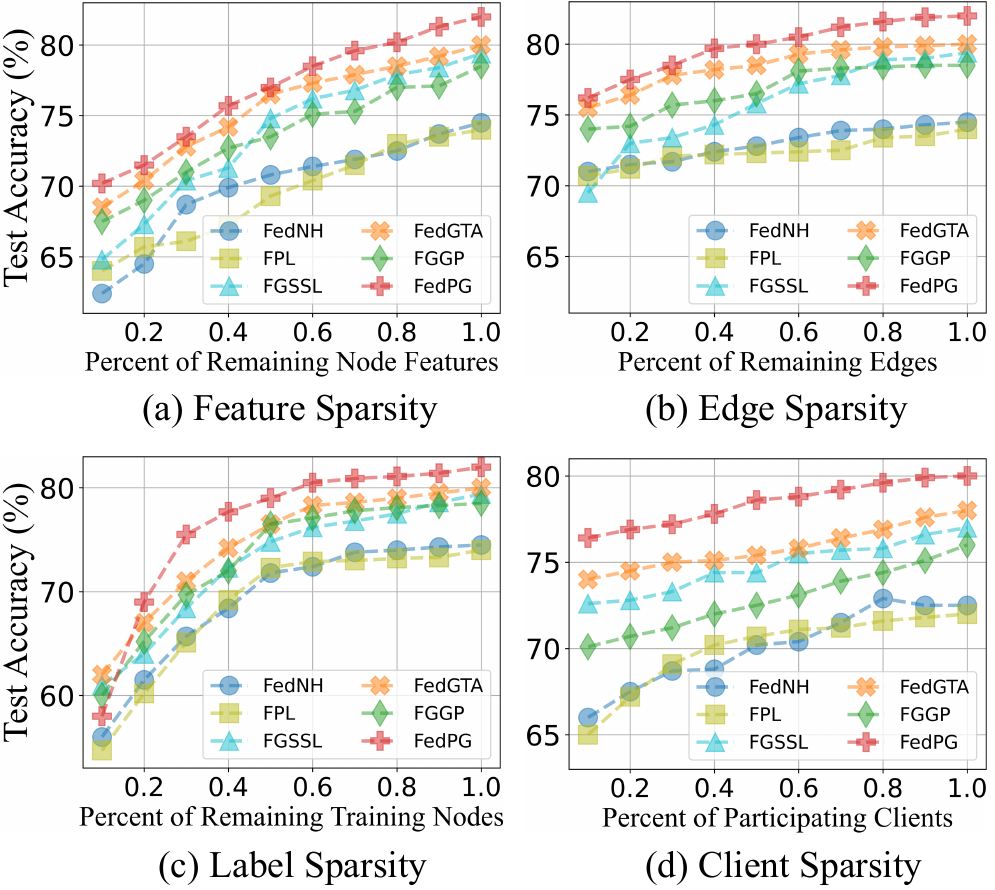}
  \caption{
    Sparsity performance on Computer with GraphSAGE.
}
  \label{fig: exp_sparsity}
\end{figure}

\vspace{0.08cm}
\subsection{Performance under Sparse Settings.}
\label{sec: Performance under Sparse Settings}
    In terms of feature sparsity, we assume the partial absence of features for unlabeled nodes. 
    To simulate edge sparsity, we randomly remove edges.
    For label sparsity, we change the ratio of labeled nodes. 
    Under feature sparsity, Fig.~\ref{fig: exp_sparsity} (a) indicates that FedGTA, FGSSL, and all FPL methods are adversely affected by low-quality local data.
    However, our proposed FedPG mitigates this issue through multi-granularity prototypes, swiftly restoring predictive performance as node features are restored.
    Such performance can also be applied to edge-sparse and label-sparse scenarios.
    Moreover, in practical FGL scenarios, it's necessary to select a subset of clients to participate in each round. 
    Hence, we conduct a 20-client split and present the convergence performance in Fig.~\ref{fig: exp_sparsity} (d). 
    According to the results, we conclude that FedPG maintains stable performance across different participation ratios.

\section{Conclusion}
    Multi-level FGL heterogeneity, as illustrated in Fig.~\ref{fig: heterogeneity_challenges}, poses as a prominent challenge to meet practical demands in distributed graph machine intelligence. 
    While some FGL methods address the DH, a general optimization framework is still lacking. 
    To address this issue, we propose a simple yet effective approach called FedPG.
    The SOTA performances coupled with low communication cost and high running efficiency demonstrate the practicality of our approach. 
    For future work, we believe it would be worthwhile to explore the integration of techniques such as graph augmentation or knowledge distillation to enable global prototypes to more effectively guide local updates.
    Additionally, investigating the effective adaptation of graph-based FPL in continual learning or contrastive learning scenarios presents a promising direction.


\vfill

\newpage
\balance{
\bibliographystyle{IEEEtran}
\bibliography{Federated_Prototype_Graph_Learning}

\begin{thebibliography}{10}
\providecommand{\url}[1]{#1}
\csname url@samestyle\endcsname
\providecommand{\newblock}{\relax}
\providecommand{\bibinfo}[2]{#2}
\providecommand{\BIBentrySTDinterwordspacing}{\spaceskip=0pt\relax}
\providecommand{\BIBentryALTinterwordstretchfactor}{4}
\providecommand{\BIBentryALTinterwordspacing}{\spaceskip=\fontdimen2\font plus
\BIBentryALTinterwordstretchfactor\fontdimen3\font minus \fontdimen4\font\relax}
\providecommand{\BIBforeignlanguage}[2]{{%
\expandafter\ifx\csname l@#1\endcsname\relax
\typeout{** WARNING: IEEEtran.bst: No hyphenation pattern has been}%
\typeout{** loaded for the language `#1'. Using the pattern for}%
\typeout{** the default language instead.}%
\else
\language=\csname l@#1\endcsname
\fi
#2}}
\providecommand{\BIBdecl}{\relax}
\BIBdecl

\bibitem{balmaseda2023app_gnn_fina1}
V.~Balmaseda, M.~Coronado, and G.~de~Cadenas-Santiagoc, ``Predicting systemic risk in financial systems using deep graph learning,'' \emph{Intelligent Systems with Applications}, p. 200240, 2023.

\bibitem{hyun2023app_gnn_fina2}
W.~Hyun, J.~Lee, and B.~Suh, ``Anti-money laundering in cryptocurrency via multi-relational graph neural network,'' in \emph{Pacific-Asia Conference on Knowledge Discovery and Data Mining}.\hskip 1em plus 0.5em minus 0.4em\relax Springer, 2023, pp. 118--130.

\bibitem{qiu2023app_gnn_fina3}
Y.~Qiu, ``Default risk assessment of internet financial enterprises based on graph neural network,'' in \emph{IEEE Information TechnoLoGy, Networking, Electronic and Automation Control Conference}, vol.~6.\hskip 1em plus 0.5em minus 0.4em\relax IEEE, 2023, pp. 592--596.

\bibitem{bang2023app_gnn_bio1}
D.~Bang, S.~Lim, S.~Lee, and S.~Kim, ``Biomedical knowledge graph learning for drug repurposing by extending guilt-by-association to multiple layers,'' \emph{Nature Communications}, vol.~14, no.~1, p. 3570, 2023.

\bibitem{qu2023app_gnn_bio2}
Z.~Qu, T.~Yao, X.~Liu, and G.~Wang, ``A graph convolutional network based on univariate neurodegeneration biomarker for alzheimer’s disease diagnosis,'' \emph{IEEE Journal of Translational Engineering in Health and Medicine}, 2023.

\bibitem{gao2023app_gnn_bio3}
Z.~Gao, H.~Ma, X.~Zhang, Y.~Wang, and Z.~Wu, ``Similarity measures-based graph co-contrastive learning for drug--disease association prediction,'' \emph{Bioinformatics}, vol.~39, no.~6, p. btad357, 2023.

\bibitem{10.1007/s00521-022-07735-y_EHGCN}
X.~Li, R.~Guo, J.~Chen, Y.~Hu, M.~Qu, and B.~Jiang, ``Effective hybrid graph and hypergraph convolution network for collaborative filtering,'' \emph{Neural Comput. Appl.}, vol.~35, no.~3, p. 2633–2646, sep 2022.

\bibitem{yang2023app_gnn_rec2}
L.~Yang, S.~Wang, Y.~Tao, J.~Sun, X.~Liu, P.~S. Yu, and T.~Wang, ``Dgrec: Graph neural network for recommendation with diversified embedding generation,'' in \emph{Proceedings of the ACM International Conference on Web Search and Data Mining, WSDM}, 2023.

\bibitem{cai2023app_gnn_rec3}
X.~Cai, C.~Huang, L.~Xia, and X.~Ren, ``Lightgcl: Simple yet effective graph contrastive learning for recommendation,'' in \emph{International Conference on Learning Representations, ICLR}, 2023.

\bibitem{wu2019sgc}
F.~Wu, A.~Souza, T.~Zhang, C.~Fifty, T.~Yu, and K.~Weinberger, ``Simplifying graph convolutional networks,'' in \emph{International Conference on Machine Learning, ICML}, 2019.

\bibitem{Hu2021ahgae}
Y.~Hu, X.~Li, Y.~Wang, Y.~Wu, Y.~Zhao, C.~Yan, J.~Yin, and Y.~Gao, ``Adaptive hypergraph auto-encoder for relational data clustering,'' \emph{IEEE Transactions on Knowledge and Data Engineering}, 2021.

\bibitem{cai2021link_prediction2}
L.~Cai, J.~Li, J.~Wang, and S.~Ji, ``Line graph neural networks for link prediction,'' \emph{IEEE Transactions on Pattern Analysis and Machine Intelligence}, 2021.

\bibitem{tan2023link_prediction4}
Q.~Tan, X.~Zhang, N.~Liu, D.~Zha, L.~Li, R.~Chen, S.-H. Choi, and X.~Hu, ``Bring your own view: Graph neural networks for link prediction with personalized subgraph selection,'' in \emph{Proceedings of the ACM International Conference on Web Search and Data Mining, WSDM}, 2023.

\bibitem{zhang2019graph_classification1}
Z.~Zhang, J.~Bu, M.~Ester, J.~Zhang, C.~Yao, Z.~Yu, and C.~Wang, ``Hierarchical graph pooling with structure learning,'' \emph{arXiv preprint arXiv:1911.05954}, 2019.

\bibitem{yang2022graph_classification3}
M.~Yang, Y.~Shen, R.~Li, H.~Qi, Q.~Zhang, and B.~Yin, ``A new perspective on the effects of spectrum in graph neural networks,'' in \emph{International Conference on Machine Learning, ICML}, 2022.

\bibitem{arya2024_TPAMI_1}
D.~Arya, D.~K. Gupta, S.~Rudinac, and M.~Worring, ``Adaptive neural message passing for inductive learning on hypergraphs,'' \emph{IEEE Transactions on Pattern Analysis and Machine Intelligence, TPAMI}, 2024.

\bibitem{fan2023_TPAMI_2}
S.~Fan, X.~Wang, C.~Shi, P.~Cui, and B.~Wang, ``Generalizing graph neural networks on out-of-distribution graphs,'' \emph{IEEE Transactions on Pattern Analysis and Machine Intelligence, TPAMI}, 2023.

\bibitem{zheng2023_TPAMI_3}
S.~Zheng, Z.~Zhu, Z.~Liu, Y.~Li, and Y.~Zhao, ``Node-oriented spectral filtering for graph neural networks,'' \emph{IEEE Transactions on Pattern Analysis and Machine Intelligence, TPAMI}, 2023.

\bibitem{qi2023_TPAMI_4}
Y.~Qi, J.~Wu, H.~Xu, and M.~Guizani, ``Blockchain data mining with graph learning: A survey,'' \emph{IEEE Transactions on Pattern Analysis and Machine Intelligence, TPAMI}, 2023.

\bibitem{yin2023_TPAMI_5}
N.~Yin, L.~Shen, H.~Xiong, B.~Gu, C.~Chen, X.-S. Hua, S.~Liu, and X.~Luo, ``Messages are never propagated alone: Collaborative hypergraph neural network for time-series forecasting,'' \emph{IEEE Transactions on Pattern Analysis and Machine Intelligence, TPAMI}, 2023.

\bibitem{pan2022fedapp_gnn_fina1}
Z.~Pan, G.~Wang, Z.~Li, L.~Chen, Y.~Bian, and Z.~Lai, ``2sfgl: A simple and robust protocol for graph-based fraud detection,'' in \emph{IEEE International Conference on Cloud Computing TechnoLoGy and Science, CloudCom}.\hskip 1em plus 0.5em minus 0.4em\relax IEEE, 2022, pp. 194--201.

\bibitem{wufederated2023fedapp_gnn_bio2}
X.~Wu, J.~Gao, M.~Bilal, F.~Dai, X.~Xu, L.~Qi, and W.~Dou, ``Federated learning-based private medical knowledge graph for epidemic surveillance in internet of things,'' \emph{Expert Systems}, p. e13372, 2023.

\bibitem{yin2022fedapp_gnn_rec1}
Y.~Yin, Y.~Li, H.~Gao, T.~Liang, and Q.~Pan, ``Fgc: Gcn based federated learning approach for trust industrial service recommendation,'' \emph{IEEE Transactions on Industrial Informatics}, 2022.

\bibitem{mai2023fedapp_gnn_rec3}
P.~Mai and Y.~Pang, ``Vertical federated graph neural network for recommender system,'' \emph{arXiv preprint arXiv:2303.05786}, 2023.

\bibitem{frasca2020sign}
F.~Frasca, E.~Rossi, D.~Eynard, B.~Chamberlain, M.~Bronstein, and F.~Monti, ``Sign: Scalable inception graph neural networks,'' \emph{arXiv preprint arXiv:2004.11198}, 2020.

\bibitem{sun2021sagn}
C.~Sun, H.~Gu, and J.~Hu, ``Scalable and adaptive graph neural networks with self-label-enhanced training,'' \emph{arXiv preprint arXiv:2104.09376}, 2021.

\bibitem{zhang2022pasca}
W.~Zhang, Y.~Shen, Z.~Lin, Y.~Li, X.~Li, W.~Ouyang, Y.~Tao, Z.~Yang, and B.~Cui, ``Pasca: A graph neural architecture search system under the scalable paradigm,'' in \emph{Proceedings of the ACM Web Conference, WWW}, 2022.

\bibitem{li2024_atp}
X.~Li, J.~Ma, Z.~Wu, D.~Su, W.~Zhang, R.-H. Li, and G.~Wang, ``Rethinking node-wise propagation for large-scale graph learning,'' in \emph{Proceedings of the ACM Web Conference, WWW}, 2024.

\bibitem{li2023fedgta}
X.~Li, Z.~Wu, W.~Zhang, Y.~Zhu, R.-H. Li, and G.~Wang, ``Fedgta: Topology-aware averaging for federated graph learning,'' \emph{Proceedings of the VLDB Endowment}, 2023.

\bibitem{zhang2024feddep}
K.~Zhang, L.~Sun, B.~Ding, S.~M. Yiu, and C.~Yang, ``Deep efficient private neighbor generation for subgraph federated learning,'' in \emph{Proceedings of SIAM International Conference on Data Mining, SDM}, 2024.

\bibitem{yao2024fedgcn}
Y.~Yao, W.~Jin, S.~Ravi, and C.~Joe-Wong, ``Fedgcn: Convergence-communication tradeoffs in federated training of graph convolutional networks,'' \emph{Advances in neural information processing systems}, vol.~36, 2023.

\bibitem{liu2023ofa}
H.~Liu, J.~Feng, L.~Kong, N.~Liang, D.~Tao, Y.~Chen, and M.~Zhang, ``One for all: Towards training one graph model for all classification tasks,'' in \emph{International Conference on Learning Representations, ICLR}, 2024.

\bibitem{li2023deepergcn}
G.~Li, C.~Xiong, G.~Qian, A.~Thabet, and B.~Ghanem, ``Deepergcn: Training deeper gcns with generalized aggregation functions,'' \emph{IEEE Transactions on Pattern Analysis and Machine Intelligence}, 2023.

\bibitem{wu2021GraphTrans}
Z.~Wu, P.~Jain, M.~Wright, A.~Mirhoseini, J.~E. Gonzalez, and I.~Stoica, ``Representing long-range context for graph neural networks with global attention,'' \emph{Advances in Neural Information Processing Systems, NeurIPS}, 2021.

\bibitem{chen2022nagphormer}
J.~Chen, K.~Gao, G.~Li, and K.~He, ``Nagphormer: A tokenized graph transformer for node classification in large graphs,'' in \emph{International Conference on Learning Representations, ICLR}, 2023.

\bibitem{kipf2016gcn}
T.~N. Kipf and M.~Welling, ``Semi-supervised classification with graph convolutional networks,'' in \emph{International Conference on Learning Representations, ICLR}, 2017.

\bibitem{hamilton2017graphsage}
W.~Hamilton, Z.~Ying, and J.~Leskovec, ``Inductive representation learning on large graphs,'' \emph{Advances in Neural Information Processing Systems, NeurIPS}, 2017.

\bibitem{mcpherson200homophily_theory1}
M.~McPherson, L.~Smith-Lovin, and J.~M. Cook, ``Birds of a feather: Homophily in social networks,'' \emph{Annual Review of SocioLoGy}, vol.~27, no.~1, pp. 415--444, 2001.

\bibitem{M2003Mixing_homophily_theory2}
M., E., J., and Newman, ``Mixing patterns in networks,'' \emph{Physical Review E}, vol.~67, no.~2, pp. 26\,126--26\,126, 2003.

\bibitem{0Networks_homophily_theory3}
O.~C. E. O.~B. Author@Ac, \emph{Networks, Crowds, and Markets}.\hskip 1em plus 0.5em minus 0.4em\relax Networks, Crowds, and Markets.

\bibitem{pei2020geomgcn}
H.~Pei, B.~Wei, K.~C.-C. Chang, Y.~Lei, and B.~Yang, ``Geom-gcn: Geometric graph convolutional networks,'' in \emph{International Conference on Learning Representations, ICLR}, 2020.

\bibitem{xie2021gcfl}
H.~Xie, J.~Ma, L.~Xiong, and C.~Yang, ``Federated graph classification over non-iid graphs,'' \emph{Advances in Neural Information Processing Systems, NeurIPS}, 2021.

\bibitem{tan2023fedstar}
Y.~Tan, Y.~Liu, G.~Long, J.~Jiang, Q.~Lu, and C.~Zhang, ``Federated learning on non-iid graphs via structural knowledge sharing,'' in \emph{Proceedings of the AAAI Conference on Artificial Intelligence, AAAI}, 2023.

\bibitem{zhang2021fedsage}
K.~Zhang, C.~Yang, X.~Li, L.~Sun, and S.~M. Yiu, ``Subgraph federated learning with missing neighbor generation,'' \emph{Advances in Neural Information Processing Systems, NeurIPS}, 2021.

\bibitem{wu2021fedgnn}
C.~Wu, F.~Wu, Y.~Cao, Y.~Huang, and X.~Xie, ``Fedgnn: Federated graph neural network for privacy-preserving recommendation,'' \emph{arXiv preprint arXiv:2102.04925}, 2021.

\bibitem{zhao2022fedgsl}
G.~Zhao, Y.~Huang, and C.~H. Tsai, ``Fedgsl: Federated graph structure learning for local subgraph augmentation,'' in \emph{IEEE International Conference on Big Data}.\hskip 1em plus 0.5em minus 0.4em\relax IEEE, 2022.

\bibitem{baek2022fedpub}
J.~Baek, W.~Jeong, J.~Jin, J.~Yoon, and S.~J. Hwang, ``Personalized subgraph federated learning,'' 2023.

\bibitem{chen2024fedgl}
C.~Chen, Z.~Xu, W.~Hu, Z.~Zheng, and J.~Zhang, ``Fedgl: Federated graph learning framework with global self-supervision,'' \emph{Information Sciences}, vol. 657, p. 119976, 2024.

\bibitem{tan2022fedproto}
Y.~Tan, G.~Long, L.~Liu, T.~Zhou, Q.~Lu, J.~Jiang, and C.~Zhang, ``Fedproto: Federated prototype learning across heterogeneous clients,'' in \emph{Proceedings of the AAAI Conference on Artificial Intelligence, AAAI}, 2022.

\bibitem{dai2023fednh}
Y.~Dai, Z.~Chen, J.~Li, S.~Heinecke, L.~Sun, and R.~Xu, ``Tackling data heterogeneity in federated learning with class prototypes,'' in \emph{Proceedings of the AAAI Conference on Artificial Intelligence, AAAI}, 2023.

\bibitem{huang2023fpl}
W.~Huang, M.~Ye, Z.~Shi, H.~Li, and B.~Du, ``Rethinking federated learning with domain shift: A prototype view,'' in \emph{IEEE/CVF Conference on Computer Vision and Pattern Recognition, CVPR}, 2023.

\bibitem{liao2023hyperfed}
X.~Liao, W.~Liu, C.~Chen, P.~Zhou, H.~Zhu, Y.~Tan, J.~Wang, and Y.~Qi, ``Hyperfed: hyperbolic prototypes exploration with consistent aggregation for non-iid data in federated learning,'' in \emph{Proceedings of the International Joint Conference on Artificial Intelligence, IJCAI}, 2023.

\bibitem{zhang2024fedtgp}
J.~Zhang, Y.~Liu, Y.~Hua, and J.~Cao, ``Fedtgp: Trainable global prototypes with adaptive-margin-enhanced contrastive learning for data and model heterogeneity in federated learning,'' in \emph{Proceedings of the AAAI Conference on Artificial Intelligence, AAAI}, 2024.

\bibitem{wan2024fggp}
G.~Wan, W.~Huang, and M.~Ye, ``Federated graph learning under domain shift with generalizable prototypes,'' in \emph{Proceedings of the AAAI Conference on Artificial Intelligence, AAAI}, 2024.

\bibitem{wang2024prototype_privacy1}
L.~Wang, Q.~Zhang, L.~Sang, Q.~Wu, and M.~Xu, ``Federated prototype-based contrastive learning for privacy-preserving cross-domain recommendation,'' \emph{arXiv preprint arXiv:2409.03294}, 2024.

\bibitem{wang2023prototype_privacy2}
R.~Wang, W.~Huang, X.~Zhang, J.~Wang, C.~Ding, and C.~Shen, ``Federated contrastive prototype learning: An efficient collaborative fault diagnosis method with data privacy,'' \emph{Knowledge-Based Systems}, vol. 281, p. 111093, 2023.

\bibitem{zuo2024prototype_privacy2}
R.~Zuo, C.~Zheng, F.~Li, L.~Zhu, and Z.~Zhang, ``Privacy-enhanced prototype-based federated cross-modal hashing for cross-modal retrieval,'' \emph{ACM Transactions on Multimedia Computing, Communications and Applications}, vol.~20, no.~9, pp. 1--19, 2024.

\bibitem{mcmahan2017fedavg}
B.~McMahan, E.~Moore, D.~Ramage, S.~Hampson, and B.~A. y~Arcas, ``Communication-efficient learning of deep networks from decentralized data,'' \emph{Artificial Intelligence and Statistics}, 2017.

\bibitem{li2020fedprox}
T.~Li, A.~K. Sahu, M.~Zaheer, M.~Sanjabi, A.~Talwalkar, and V.~Smith, ``Federated optimization in heterogeneous networks,'' \emph{Proceedings of Machine Learning and Systems, MLSys}, 2020.

\bibitem{karimireddy2020scaffold}
S.~P. Karimireddy, S.~Kale, M.~Mohri, S.~Reddi, S.~Stich, and A.~T. Suresh, ``Scaffold: Stochastic controlled averaging for federated learning,'' in \emph{International Conference on Machine Learning, ICML}, 2020.

\bibitem{li2021moon}
Q.~Li, B.~He, and D.~Song, ``Model-contrastive federated learning,'' in \emph{Proceedings of the IEEE/CVF Conference on Computer Vision and Pattern Recognition, CVPR}, 2021.

\bibitem{gao2022feddc}
L.~Gao, H.~Fu, L.~Li, Y.~Chen, M.~Xu, and C.-Z. Xu, ``Feddc: Federated learning with non-iid data via local drift decoupling and correction,'' in \emph{Proceedings of the IEEE/CVF Conference on Computer Vision and Pattern Recognition, CVPR}, 2022.

\bibitem{zhang2024fedgt}
Z.~Zhang, Q.~Hu, Y.~Yu, W.~Gao, and Q.~Liu, ``Fedgt: Federated node classification with scalable graph transformer,'' \emph{arXiv preprint arXiv:2401.15203}, 2024.

\bibitem{xie2023fedlit}
H.~Xie, L.~Xiong, and C.~Yang, ``Federated node classification over graphs with latent link-type heterogeneity,'' in \emph{Proceedings of the ACM Web Conference, WWW}, 2023.

\bibitem{huang2023fgssl}
W.~Huang, G.~Wan, M.~Ye, and B.~Du, ``Federated graph semantic and structural learning,'' in \emph{Proceedings of the International Joint Conference on Artificial Intelligence, IJCAI}, 2023.

\bibitem{guo2023GCFGAE}
K.~Guo, Y.~Fang, Q.~Huang, Y.~Liang, Z.~Zhang, W.~He, L.~Yang, K.~Chen, X.~Liu, and W.~Guo, ``Globally consistent federated graph autoencoder for non-iid graphs,'' in \emph{Proceedings of the International Joint Conference on Artificial Intelligence, IJCAI}, 2023.

\bibitem{li2024adafgl}
X.~Li, Z.~Wu, W.~Zhang, H.~Sun, R.-H. Li, and G.~Wang, ``Adafgl: A new paradigm for federated node classification with topology heterogeneity,'' \emph{arXiv preprint arXiv:2401.11750}, 2024.

\bibitem{xu2018jknet}
K.~Xu, C.~Li, Y.~Tian, T.~Sonobe, K.-i. Kawarabayashi, and S.~Jegelka, ``Representation learning on graphs with jumping knowledge networks,'' in \emph{International Conference on Machine Learning, ICML}, 2018.

\bibitem{chen2020gbp}
M.~Chen, Z.~Wei, B.~Ding, Y.~Li, Y.~Yuan, X.~Du, and J.-R. Wen, ``Scalable graph neural networks via bidirectional propagation,'' \emph{Advances in Neural Information Processing Systems, NeurIPS}, 2020.

\bibitem{zhu2024fedtad}
Y.~Zhu, X.~Li, Z.~Wu, D.~Wu, M.~Hu, and R.-H. Li, ``Fedtad: Topology-aware data-free knowledge distillation for subgraph federated learning,'' \emph{arXiv preprint arXiv:2404.14061}, 2024.

\bibitem{zhu2021Non-iid_fl_survey1}
H.~Zhu, J.~Xu, S.~Liu, and Y.~Jin, ``Federated learning on non-iid data: A survey,'' \emph{Neurocomputing}, vol. 465, pp. 371--390, 2021.

\bibitem{matsuda2022Non-iid_fl_survey2}
K.~Matsuda, Y.~Sasaki, C.~Xiao, and M.~Onizuka, ``An empirical study of personalized federated learning,'' \emph{arXiv preprint arXiv:2206.13190}, 2022.

\bibitem{li2022Non-iid_fl_survey3}
Q.~Li, Y.~Diao, Q.~Chen, and B.~He, ``Federated learning on non-iid data silos: An experimental study,'' in \emph{International Conference on Data Engineering, ICDE}, 2022.

\bibitem{blondel2008louvain}
V.~D. Blondel, J.-L. Guillaume, R.~Lambiotte, and E.~Lefebvre, ``Fast unfolding of communities in large networks,'' \emph{Journal of Statistical Mechanics: Theory and Experiment}, vol. 2008, no.~10, p. P10008, 2008.

\bibitem{karypis1998metis}
G.~Karypis and V.~Kumar, ``A fast and high quality multilevel scheme for partitioning irregular graphs,'' \emph{SIAM Journal on Scientific Computing}, vol.~20, no.~1, pp. 359--392, 1998.

\bibitem{jeong2018feddistill}
E.~Jeong, S.~Oh, H.~Kim, J.~Park, M.~Bennis, and S.-L. Kim, ``Communication-efficient on-device machine learning: Federated distillation and augmentation under non-iid private data,'' \emph{arXiv preprint arXiv:1811.11479}, 2018.

\bibitem{zhu2021fedgen}
Z.~Zhu, J.~Hong, and J.~Zhou, ``Data-free knowledge distillation for heterogeneous federated learning,'' in \emph{International Conference on Machine Learning, ICML}.\hskip 1em plus 0.5em minus 0.4em\relax PMLR, 2021.

\bibitem{wu2022fedkd}
C.~Wu, F.~Wu, L.~Lyu, Y.~Huang, and X.~Xie, ``Communication-efficient federated learning via knowledge distillation,'' \emph{Nature communications}, vol.~13, no.~1, p. 2032, 2022.

\bibitem{xu2018gin}
K.~Xu, W.~Hu, J.~Leskovec, and S.~Jegelka, ``How powerful are graph neural networks?'' 2019.

\bibitem{chen2020gcnii}
M.~Chen, Z.~Wei, Z.~Huang, B.~Ding, and Y.~Li, ``Simple and deep graph convolutional networks,'' in \emph{International Conference on Machine Learning, ICML}, 2020.

\bibitem{akiba2019optuna}
T.~Akiba, S.~Sano, T.~Yanase, T.~Ohta, and M.~Koyama, ``Optuna: A next-generation hyperparameter optimization framework,'' in \emph{Proceedings of the ACM SIGKDD Conference on Knowledge Discovery and Data Mining, KDD}, 2019.

\end{thebibliography}
}

\end{document}